\documentclass{article}

\usepackage{arxiv}

\usepackage{xspace}

\usepackage{hyperref}
\usepackage{algorithm}
\usepackage{algpseudocode}
\usepackage{bm}
\usepackage{makecell}
\usepackage{multirow}
\usepackage{subfigure}
\usepackage{todonotes}
\usepackage[framemethod=tikz]{mdframed}
\usepackage{adjustbox}
\usepackage{framed}
\usepackage{amsmath} 
\usepackage{amssymb} 
\usepackage{enumitem}
\usepackage{booktabs}
\DeclareMathOperator*{\argmin}{arg\,min}

\newcommand{\themodel}{\ensuremath{\mathrm{ARES}}\xspace}
\newcommand{\GNN}{\ensuremath{\mathrm{GNN}}\xspace}
\newcommand{\HST}{\ensuremath{\mathrm{HST}}\xspace}
\newcommand{\score}{\ensuremath{\mathrm{score}}\xspace}
\newcommand{\Cache}{\ensuremath{\mathrm{Cache}}\xspace}

\title{ARES: Anomaly Recognition Model For Edge Streams}

\author{
\href{https://orcid.org/0000-0002-0961-4151}{\includegraphics[scale=0.06]{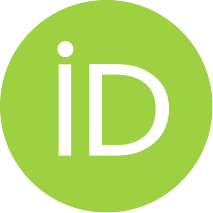}\hspace{1mm}Simone Mungari} \\
	University of Calabria\\
        ICAR-CNR\\
        Revelis s.r.l.\\
	\texttt{simone.mungari@unical.it} \\
    \And
    \href{https://orcid.org/0000-0002-8339-7773}{\includegraphics[scale=0.06]{orcid.pdf}\hspace{1mm}Albert Bifet} \\
	Artificial Intelligence Institute, University of Waikato, NZ\\
    LTCI, Télécom Paris, Institut Polytechnique de Paris\\
	\texttt{abifet@waikato.ac.nz} \\
    \And 
    \href{https://orcid.org/0000-0001-9672-3833}{\includegraphics[scale=0.06]{orcid.pdf}\hspace{1mm}Giuseppe Manco} \\
	ICAR-CNR\\
	\texttt{giuseppe.manco@icar.cnr.it}
    \And 
    \href{https://orcid.org/0000-0002-3732-5787}{\includegraphics[scale=0.06]{orcid.pdf}\hspace{1mm}Bernhard Pfahringer} \\
	Artificial Intelligence Institute, University of Waikato, NZ\\
	\texttt{bernhard@waikato.ac.nz}
   } 
\begin{document}

\maketitle

\begin{abstract}
Many real-world scenarios involving streaming information can be represented as temporal graphs, where data flows through dynamic changes in edges over time. Anomaly detection in this context has the objective of identifying unusual temporal connections within the graph structure. Detecting edge anomalies in real time is crucial for mitigating potential risks. Unlike traditional anomaly detection, this task is particularly challenging due to concept drifts, large data volumes, and the need for real-time response. 
To face these challenges, we introduce \themodel, an unsupervised anomaly detection framework for edge streams. \themodel combines Graph Neural Networks (GNNs) for feature extraction with Half-Space Trees (HST) for anomaly scoring. GNNs capture both spike and burst anomalous behaviors within streams by embedding node and edge properties in a latent space, while HST partitions this space to isolate anomalies efficiently. \themodel operates in an unsupervised way without the need for prior data labeling. To further validate its detection capabilities, we additionally incorporate a simple yet effective supervised thresholding mechanism. This approach leverages statistical dispersion among anomaly scores to determine the optimal threshold using a minimal set of labeled data, ensuring adaptability across different domains. We validate \themodel through extensive evaluations across several real-world cyber-attack scenarios, comparing its performance against existing methods while analyzing its space and time complexity. The code used to perform the experiments is publicly available at \url{https://github.com/AnomalyRecognitionModelForEdgeStreams/ARES}.
\end{abstract}


\maketitle

\section{Introduction}
\label{sec:introduction}
In the context of cyber-security, anomaly detection is one of the most important tasks, whose aim is to identify potential cyber-attacks \cite{DBLP:journals/jnca/XieHTP11, DBLP:conf/sac/Mahoney03} such as SQL Injections, Botnets, Distributed Denial of Service (DDoS) attacks \cite{DBLP:journals/jnca/AhmedMH16, DBLP:conf/asunam/PapalexakisBS12}, as well as financial frauds \cite{DBLP:journals/fgcs/AhmedM016}. Most of these threats can be represented as temporal connections, i.e., temporal edges, between endpoints. Automatically recognizing anomalous edges in data streaming domains is hence fundamental for the early detection of noxious activities.

\begin{figure}
    \begin{center}
    \subfigure[T1]{
        \includegraphics[width=.2\textwidth]{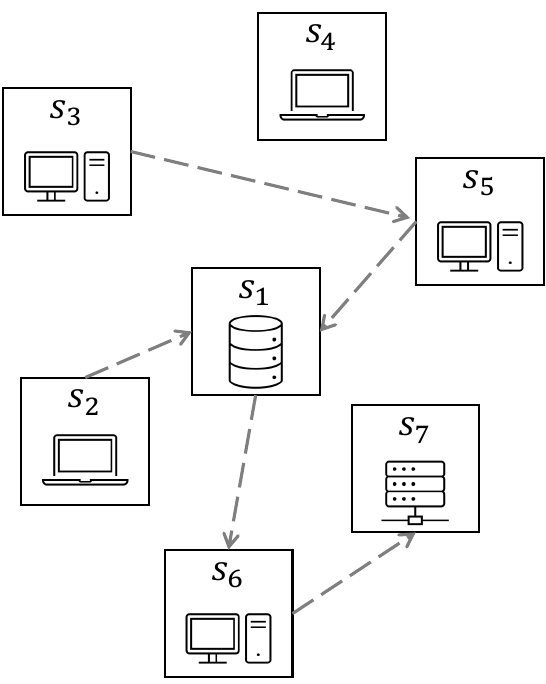}
    }
    \subfigure[T2]{
        \includegraphics[width=.2\textwidth]{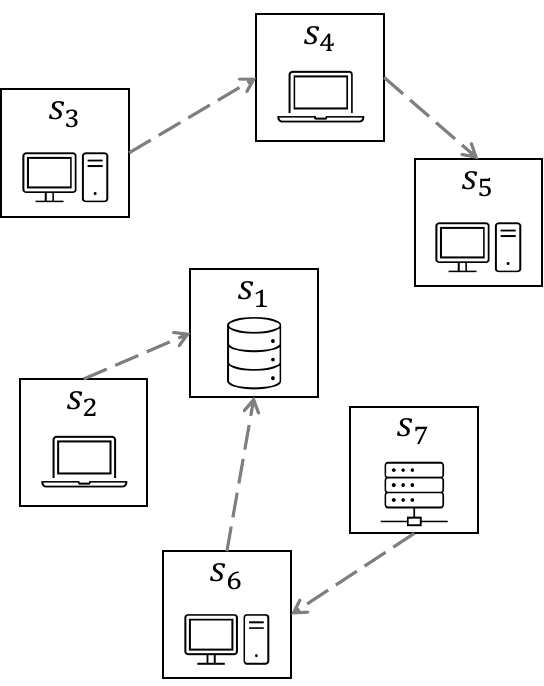}
    }
    \subfigure[T3]{
        \includegraphics[width=.2\textwidth]{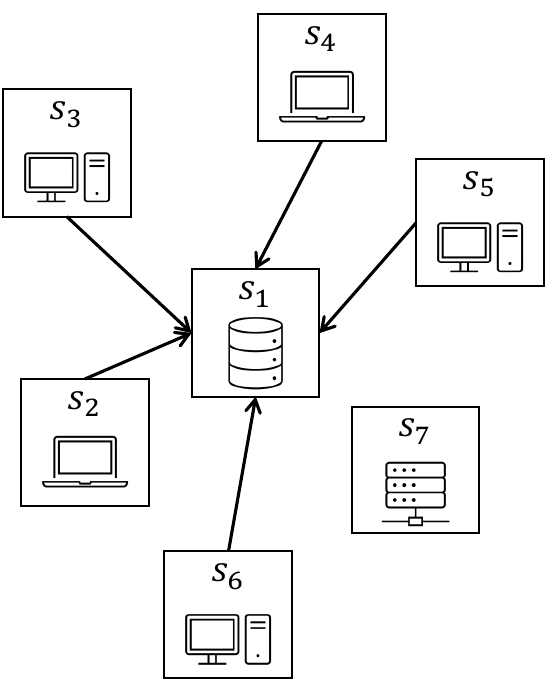}
    }
    \caption{Connections among endpoints over time (T1, T2, T3). Fig (a) and (b) depict the typical behavior of connections among endpoints. Fig (c) shows a significant increase of connections from $s_2, ..., s_6$ to $s_1$.}
    \label{fig:usecase}
    \end{center}
\end{figure}

Edge anomalies can be generally categorized into ``burst'' or ``spike'' (which can be further divided in more specific sub-categories \cite{DBLP:conf/wsdm/Chang0SASL21}), based on whether they involve multiple nodes and connections or represent an anomaly relative to a specific connection. A noteworthy example of burst anomaly is the classical DDoS scenario, described in Figure \ref{fig:usecase}, where a peer ($s_1$, representing a hosting website) is targeted by other nodes in a network, who massively connect to the peer within a specific timeframe. Early recognition of such abnormal connections is crucial for avoiding potential cyber-attacks and ensuring a reliable quality of service.


Discovering potential harmful activity within the flow of such temporal interactions poses several challenges. 
First, the streaming nature of the underlying data requires the capability of dealing with changes in the underlying edge distributions (described as concept drifts in the literature) or the rise of new anomalies. 
Additionally, data streaming scenarios are inherently hard to cope with due to the massive amount of data to manage, and by contrast they require the capability of classifying edges in real-time (or near real-time).

The recent literature has paid significant attention to the challenge of identifying anomalies in edge streams. Supervised approaches, such as those in \cite{DBLP:journals/ijis/ZhangLWSD22, DBLP:journals/compsec/ZolaSBGU22}, face two main limitations. First, they are restricted to detecting previously known anomalies. Second, they rely on labeled data, which can be difficult to obtain in real-world operational scenarios. In contrast, semi-supervised and unsupervised algorithms \cite{DBLP:conf/ijcai/ZhengLLLG19, DBLP:journals/tkdd/LiuTZ12, DBLP:journals/tkde/XuPWW23, DBLP:conf/aaai/0001HYSF20, DBLP:journals/tkdd/BhatiaLHYSF22, DBLP:conf/kdd/0001WKSYH23, DBLP:conf/sdm/RanshousHSS16, DBLP:conf/icdm/EswaranF18, DBLP:conf/wsdm/Chang0SASL21, DBLP:conf/kdd/BelthZK20, DBLP:conf/kdd/0001WKSYH23, DBLP:conf/kdd/LeeKS24}  address these limitations by reducing the need for labeled examples.
The approaches in \cite{DBLP:conf/sdm/RanshousHSS16, DBLP:conf/icdm/EswaranF18} focus on spike anomalies and tend to overlook anomalies originated by coordinated behavior involving multiple edges. On the other hand, other methods focus on burstiness \cite{DBLP:conf/wsdm/Chang0SASL21, DBLP:conf/kdd/BelthZK20, DBLP:journals/tkdd/BhatiaLHYSF22, DBLP:conf/kdd/0001WKSYH23} and disregard anomalies involving sparsely-connected graph components.
%
Another significant challenge with most unsupervised methods is that they do not provide a clear distinction between normal and anomalous edges. Instead, they generate a range of anomaly scores that must be compared to domain-specific thresholds to assess how abnormal the data is. The choice of these thresholds is critical, as even small adjustments can have a major impact on detection performance. 

In this paper, we introduce \themodel (Anomaly Recognition Model for Edge Streams), an unsupervised anomaly detection model for edge streams. 
First, \themodel automatically extracts and maps node and edge features from the stream to a latent space. To do so, the model employs a Graph Neural Network (GNN), which is the state-of-the-art approach for graph representation learning tasks. Second, for assigning an anomaly score for each edge, we endow \themodel with a tree-based model named Half-Space Trees (HST). As such, \themodel randomly divides the latent space and separates normal from suspicious edges, according to their representations.
The degree of anomaly in a component is revealed through the geometrical embeddings, 
enabling the representation of both spiking and bursty behaviors in terms of geometrical proximity. HST further enhances the model by efficiently exploring the latent space and isolating anomalies through random partitioning, thus  enabling the detection of a broad range of anomaly types.
Our framework integrates a simple GNN to minimize computational overhead while delegating stream processing to the HST. This design ensures efficient anomaly detection by balancing the strengths of both components—leveraging the GNN for feature extraction and the HST for handling dynamic edge streams. More importantly, we provide both theoretical and empirical evidence that \themodel offers an efficient processing of the data stream, as it complexity is only affected by the computational complexity of the GNN component.

As an additional aspect that we investigate, we further propose a simple yet effective supervised thresholding mechanism that can be combined with the \themodel scoring framework. Specifically, we propose a domain-agnostic method that leverages the Gini index—a measure of statistical dispersion among anomaly scores—to strongly separate true anomalies effectively. Although the method requires a minimal amount of supervision, it intrinsically explores the distribution of the anomaly scores against true labels, thus identifying the optimal cut-off in an effective way.


We empirically demonstrate the capabilities of \themodel by performing an extensive evaluation on seven different real-world scenarios, which represent different types of cyber-attacks. Our analysis also includes a comparison with existing methods and considers theoretical aspects such as space and time complexities.
To summarize, our contributions are the following.
\begin{itemize}[leftmargin=*]
    \item We devise \themodel, a simple yet effective model for identifying anomalies in edge streaming data that combines Graph Neural Networks and Half-Space Trees.
    \item We show both theoretically and empirically that \themodel is amenable for efficient edge stream processing and can effectively generate meaningful node and edge embeddings that consistently uncover edge anomalies. 
    \item We additionally equip \themodel with an algorithmic validation-based supervised method for choosing the best threshold, which can be used in real-world scenarios for discriminating between normal and abnormal activities. 
    \item Through an extensive evaluation, we show that \themodel\ outperforms SOTA methods using seven different real-world scenarios. 
\end{itemize}

The rest of the paper is organized as follows. Section~\ref{sec:related_work} discusses the contributions of the current literature in the area of graph anomaly detection. Section~\ref{sec:approach} formalizes the problem and provides a technical description of the proposed solution.
Theoretical insights and model properties are discussed in  Section~\ref{sec:theoretical}. The experimental evaluation is described in Section~\ref{sec:experiments}.
Finally, Section~\ref{sec:conclusions} concludes the paper and set pointers for further research.

\section{Related work}
\label{sec:related_work}

Our contribution is placed in the general area of Anomaly Detection. Our focus is on anomalies occuring within graph representations and their evolution in a streaming scenario. In this perspective, we identify two main topics related to our work, namely graph/node anomalies and edge anomalies.

\paragraph{\textbf{Graph and Node Anomaly Detection.}}
The problem of detecting anomalous nodes, or even anomalous graph configuration, has been extensively studied in the current literature, given the importance of accounting for inter-dependencies during the anomaly detection process, which can be also exploited for describing and explaining the inherent anomalies. Approaches to graph anomaly detection focus on both static and dynamic aspects~\cite{https://doi.org/10.1002/wics.1347}, and consider structure, local and global features~\cite{10.1007/s10618-014-0365-y}. Graph iForest~\cite{DBLP:conf/ijcnn/ZambonLA22} adapts the Isolation Forest algorithm for identifying anomalous graphs in static scenarios. Isolation Forest (IF) \cite{DBLP:journals/tkdd/LiuTZ12} is a popular tree-based method for anomaly detection on tabular data. It isolates anomalies by dividing the feature space into sub-spaces. Particular emphasis is also given by explainability and interpretability~\cite{NEURIPS2023_1c6f0686}. 

Recent studies employ advanced techniques based on Neural Networks for discovering anomalies. Deep representation learning techniques provide expressive representations facilitating the task of dividing anomalies from normal data \cite{DBLP:journals/tkde/MaWXYZSXA23}, at the expense of interpretability. \cite{DBLP:journals/tkde/XuPWW23} combines Isolation Forests with Neural Networks (NNs). In particular, the author propose to initialize the weights of an ensemble of NNs, exploiting Isolation Forest to identify anomalous data points based on the embeddings produced by the former. Graph Neural Networks~\cite{9046288} have recently gained particular attention~\cite{pmlr-v162-tang22b,8862186,9906987,DBLP:journals/tkde/MaWXYZSXA23}.  The authors of \cite{DBLP:journals/ijis/ZhangLWSD22} present a framework for detecting malwares, exploiting both Graph Neural Networks and Recurrent Neural Networks in order to capture temporal and structural correlations. \cite{DBLP:journals/compsec/ZolaSBGU22} focuses on the detection the anomalous nodes which are the source of potential harmful connections within a network. 

\paragraph{\textbf{Edge Anomaly Detection.}}
The detection of anomalous edges in both static and dynamic scenarios seeks to uncover unexpected connections at different levels: locally, where a single edge between two endpoints is deemed anomalous, and globally, where multiple connections within a group of nodes raise suspicion.
\cite{DBLP:conf/kdd/YuCAZCW18} introduced NetWalk, which detects node and edge anomalies in temporal networks by clustering their corresponding embeddings. The anomaly score for a data point (edge or node) is determined by its proximity to the nearest cluster center. AddGraph~\cite{DBLP:conf/ijcai/ZhengLLLG19} combines Graph Convolutional Network (GCN) with attention methods and employs a semi-supervised training procedure to cope with the scarcity of labelled anomalies.

In the edge streaming context, PENminer~\cite{DBLP:conf/kdd/BelthZK20} is a framework for studying persistent patterns in evolving networks by extending the concept of temporal motifs, i.e., recurrent patterns in temporal edges.  F-Fade~\cite{DBLP:conf/wsdm/Chang0SASL21} models the distribution of edge frequency over time and identifies anomalies by assessing the likelihood of the observed frequency of incoming interactions. In a similar vein, SedanSpot~\cite{DBLP:conf/icdm/EswaranF18} introduces a PageRank-like algorithm that exploits the Marginal Proximity Increase metric to compute the anomaly score, which is based on edge frequencies. SLADE-H~\cite{DBLP:conf/kdd/LeeKS24} extend the Temporal Graph Networks architecture proposed by \cite{tgn_icml_grl2020} for detecting anomalous edges, employing node memories for tracking the graph evolution along the stream. The proposed model assigns anomaly scores by measuring the cosine similarity between the node’s current and previous memory vectors. MIDAS~\cite{DBLP:conf/aaai/0001HYSF20, DBLP:journals/tkdd/BhatiaLHYSF22} exploits count-min sketches (CMS, specific data structures tailored for counting node degrees) to identify anomalies. AnoGraph~\cite{DBLP:conf/kdd/0001WKSYH23} employs an enhanced version of CMS  (high-order count-min sketches, H-CMS). These approaches are particularly effective in detecting global (bursting) features and instead may struggle in scenarios where anomalies are not related to edge frequencies. 

Compared to the aforementioned approaches, our approach, similarly to \cite{DBLP:conf/kdd/LeeKS24}, relies on the capability of GNNs to automatically extract topological and semantic features of the underlying edge stream. This allows us to boost the detection both at a local and a global level, as explained in the next sections. 




\section{Anomaly Recognition for Edge Streams}
\label{sec:approach}

We begin our treatment by fixing some notation and setting the main problem statement.
Let $\mathcal{G} = \{e_1, e_2, ...,\}$ be a (potentially infinite) stream of  edges representing connections between entities. We denote as $V_t$ and $E_t$ the set of all the nodes and edges comprised within the stream until time $t$, respectively. 
Each $e_t$ is a tuple in the form $e_t=(s, d, t)$ which occurs at time $t$, and where $s, d \in V_t$ are the source and the destination node, respectively.
%
%
Let $X_t \in \mathbb{R}^{|V_t| \times K}$ be the feature matrix associated with nodes, and $G_t=\langle V_t, E_t, X_t\rangle$ be the temporal directed multigraph representing the evolution within  $\mathcal{G}$ up to time $t$. 
We assume that nodes exhibit at least one feature. Scenarios where no explicit features are available are modeled by considering the node identifiers as the only available feature. This modeling choice is crucial for devising meaningful representations of structural properties, as explained in the following section. 


Each edge $e_t$ can be characterized by a binary label $l^{e_t} \in \{0,1\}$. In particular, $l^{e_t}=1$ denotes that $e_t$ can be considered anomalous, whereas $l^{e_t}=0$ denotes normality. 

\begin{figure}
    \centering
    \includegraphics[width=0.7\textwidth]{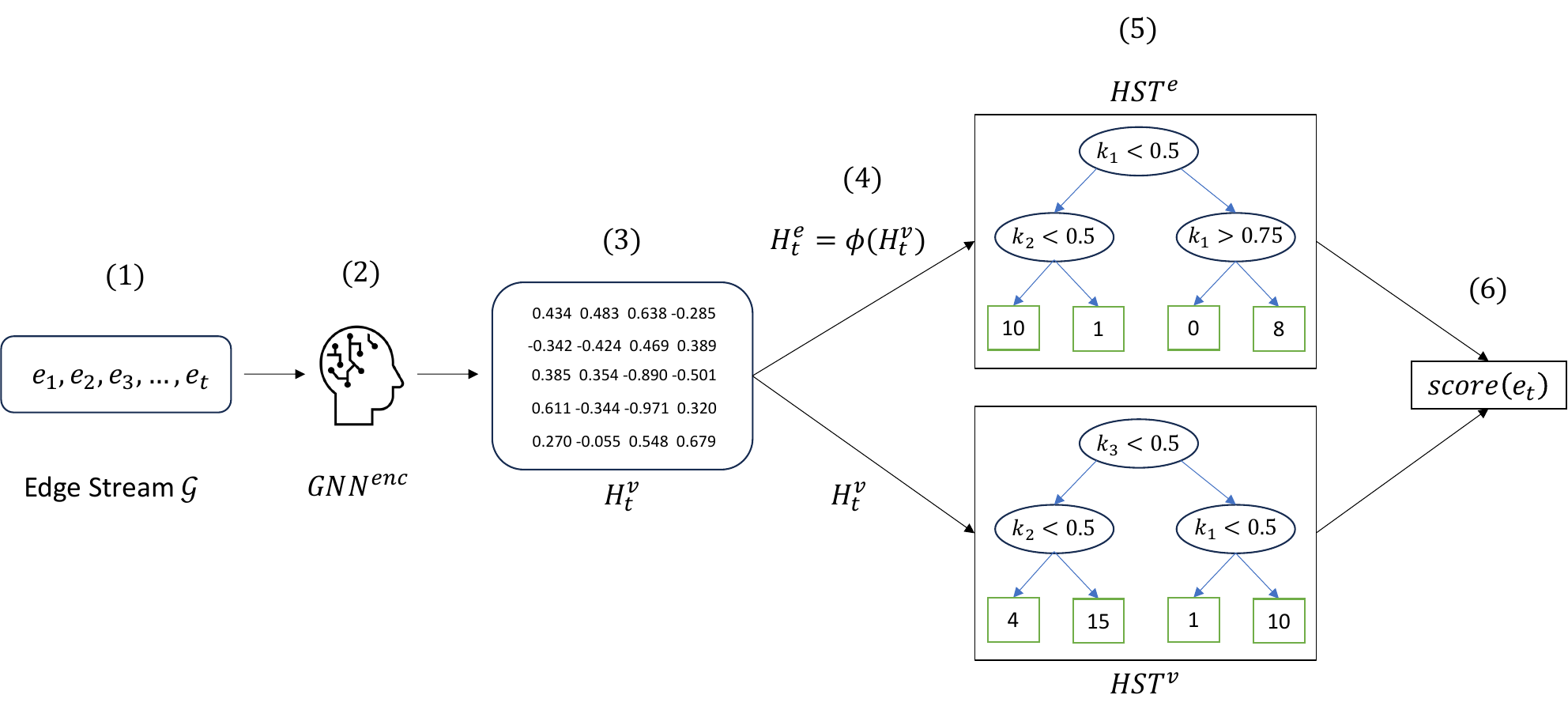}
    
    \caption{Framework overview. 
    Given a streaming of edges denoted as $\mathcal{G}$ (1), we use a pre-trained Graph Neural Network, $GNN^{enc}$ (2), to derive a set of node embeddings, represented as $H_{t}^v$ (3). We also generate a set of edge embeddings denoted as $H_{t}^e$ (4). These embeddings serve as inputs to two Half-Space Trees, designated as $HST^v$ and $HST^e$, which encode unexpectedness for both nodes and edges according to their positioning in the latent space (5). The final anomaly score (6) is hence computed by combining information from these trees both at a node and an edge level.
    }
    \label{fig:framework}
\end{figure}

Our objective is to estimate the probability $P(l^{e_t} = 1|e_t, G_{t})$, given the edge $e_t \in \mathcal{G}$ and the temporal graph $G_t=\langle V_t, E_t, X_t\rangle$ which encodes the evolution of $\mathcal{G}$. In practice, the research question we are interested in addressing is: how likely is it to consider $e_t \in \mathcal{G}$ anomalous, given its features and the current history $G_{t}$? 
To tackle this problem, we leverage GNN embeddings and Half-Space Trees. Figure \ref{fig:framework} represents a graphical overview of the framework. The core idea is to utilize a Half-Space Tree model to distinguish anomalous edges from normal ones in $\mathcal{G}$. By assuming that elements to be analyzed can be mapped into an appropriate representation space, Half-Space Trees provide an efficient partitioning of such space, thus enabling a dynamic yet fast and accurate identification of isolated data points. To ensure a semantically consistent representation of the evolving graph components, we use a Graph Neural Network to generate embeddings for both nodes and edges.
Thus, the two main components of our framework are:
\begin{enumerate}[leftmargin=*]
    \item \textit{Embedding generator}, which maps edges and nodes into a latent space. 
   \item \textit{Anomaly scorer}, which computes anomaly scores for edges $e_t = (s, d, t)$ by leveraging the latent representation of the edge ($e_t$), along with the embeddings of the source ($s$) and destination ($d$) nodes.
\end{enumerate}

\subsection{Embedding generator}
To map edges and nodes to a latent space by generating meaningful embeddings that translate structural properties into spatial features, we utilize Graph Neural Networks, which have been proven effective for learning tasks involving graphs~\cite{9046288}. 
In particular, we devise a Graph Autoencoder (GAE) \cite{DBLP:conf/cikm/WangPLZJ17, DBLP:journals/corr/KipfW16a, DBLP:conf/complexnetworks/RidaAP21, DBLP:conf/iscc/ZhuML20} architecture for learning the latent representations of nodes. A graph autoencoder is an encoder-decoder model $\tilde{G} = D(E(G))$, where $G$ is a graph and $E(G) \in \mathbb{R}^{n \times d}$ is the encoding of all nodes $V\in G$. It compresses the original data into latent vectors ($E(G)$) and reconstructs the original data from these vectors ($D(E(G))$), such that $G\approx \tilde{G}$. Both the encoder and decoder are GNNs. We denote them as $E \equiv \GNN^{enc}$ and $D \equiv \GNN^{dec}$, respectively. For our purposes, we adopt GraphSAGE~\cite{hamilton2018inductive} to define the architecture of the GNN components.

After training the GAE, we can generate node embeddings using only the encoder component $\GNN^{enc}$.
Specifically, we compute $h^{v}_t \in \mathbb{R}^{d}$, the embedding of the node $v$ at time $t$, and $h^{e}_t \in \mathbb{R}^{d}$, which represents the embedding of the edge $e$. The embedding $h^{v}_t$ is obtained directly from $\GNN^{enc}$. The edge embedding $h^{e}_t$ is obtained by combining the embeddings of the source and destination nodes. For this, we adopt two alternative strategies, inspired by prior work~\cite{DBLP:conf/kdd/GroverL16}. Given an edge $e_t=(s, d, t)$ at time $t$, the first alternative consists in averaging the embeddings of both the source and destination nodes, 
\begin{equation}\label{eq:edgemean}
    h_t^{e_t} = \phi_1(h_t^{s}, h_t^{d}) =  \frac{h_t^{s} + h_t^{d}}{2}
\end{equation}
Alternatively, we can compute the difference between the node embeddings, capturing directional information through the asymmetry of this difference:
\begin{equation}\label{eq:edgesub}
    h_t^{e_t} = \phi_2(h_t^{s}, h_t^{d}) = h_t^{d} - h_t^{s}
\end{equation}

\subsection{Anomaly Scoring}

To efficiently explore the resulting latent space, we adopt Half-Space Trees (HST)\cite{DBLP:conf/ijcai/TanTL11}, a robust tree ensemble framework designed for anomaly detection in streaming environments.
HST belongs to the broader family of isolation-based anomaly detection methods. We chose HST for anomaly scoring due to its demonstrated effectiveness and reliability in streaming data scenarios. Like Isolation Forest\cite{DBLP:journals/tkdd/LiuTZ12}, HST identifies anomalies by recursively partitioning the data. However, unlike Isolation Forest, HST is specifically tailored for real-time applications. One of its key strengths is the use of a sliding window mechanism, which enables the model to adapt dynamically to concept drift and evolving data distributions. Another advantage is that HST does not require a dedicated training phase, in contrast to many other isolation-based techniques~\cite{samariya2023comprehensive}, which helps reduce computational overhead.
Overall, isolation-based methods like HST offer several benefits: (i) low computational and memory requirements, (ii) scalability to large datasets, and (iii) the ability to detect both global and local anomalies~\cite{samariya2023comprehensive}.

Technically, HST builds each tree iteratively by randomly choosing a feature and a splitting value to partition the space. Each leaf node in the tree keeps a count of the number of data instances that fall within its corresponding region. Data points that appear in sparsely populated subregions are then considered anomalous.
More precisely, for a generic input $X$, HST computes the anomaly score by combining the anomaly scores exhibited by each tree in the ensemble. Each score considers both the sparsity and the extention of the subregion where $X$ occurs: 
\begin{equation}
    \HST(X) = \frac{1}{Z} \mathrm{Node}.r \times 2^{\mathrm{Node}.h}
\end{equation}
Here, $\mathrm{Node}.r$ is the counter and $\mathrm{Node}.h$ is the depth level of the leaf node where $X$ is located (i.e., the number of inequality constraints that border the subspace). $Z$ is a normalization constant. 
The counts within leaf nodes are updated dynamically as data streams in, using a window-based strategy to track changes over time. 

Within \themodel, we exploit two HST instances, namely \emph{$\HST^v$} and \emph{$\HST^e$}, to combine anomalies both at node and edge levels. The final score for a given edge $e_t=(s,d,t)$ is then obtained through a weighted average:
\begin{equation}\label{eq:anomalyscore}
\begin{split}
        \score(e_t) = & w_s \HST^v(h_{t}^{s}) + w_d \HST^v(h_{t}^{d}) \\
        & + w_e \HST^e(h_{t}^{e_t})
\end{split}
\end{equation}
The (normalized) weights $w_1$, $w_2$, and $w_3$ reflect the contributions of the source, destination node and edge embeddings to the overall anomaly score. Within our framework, $\score(e_t)$ encodes the probability $P(l^{e_t}=1|e_t, G_{t})$. 

\subsection{Processing Edge Streams}
For a given stream $\mathcal{G}$, the framework starts by training the GAE on an initial snapshot $G_t$ of the edge stream. Next, we freeze the neural network's weights and exploit $\GNN^{enc}$ for the embeddings throughout $\mathcal{G}$. For this, we build and initialize the internal data structures of both $\HST^v$ and $\HST^e$, using a subset of the data used for training the GAE. 

Algorithm~\ref{alg:streaming} shows the main pipeline of the framework. The algorithm receives as input the pre-trained $\GNN^{enc}$, initialized HSTs, and the data stream $\mathcal{G}$. The main flow hence consists in incrementally considering new edges $e_t \in \mathcal{G}$ (lines 5-18). For each (batch of) new edges, the current graph $G_{t}$ selected, (line 6) and the $\GNN^{enc}$ component generates the embeddings for the source and destination nodes ($s,d \in e_t$) (line 10). These embeddings are hence combined (line 11) to generate the embedding $h^{e_t}_t$ (through either Equation~\ref{eq:edgemean} or Equation~\ref{eq:edgesub}). The anomaly score $score^{e_t}$ is hence computed (line 12). 

An important consideration concerning the embedding is that restricting the training of $\GNN^{enc}$ on only the initial snapshot $G_t$ of the edge stream does not inherently limit the representational power of the GNN, a design choice also adopted in prior approaches~\cite{DBLP:journals/ipm/CoppolilloMRFMBM25}.
This is because the underlying temporal graph $G_t$ is continuously updated as new edges arrive. As the graph evolves, it dynamically integrates new structural and relational information, allowing node and edge embeddings to reflect the latest changes. Consequently, even without retraining, the model remains capable of capturing emerging patterns and dependencies within the graph.

To compute anomaly scores, we introduce two variants within our framework: \themodel-Static and \themodel-Dynamic. In \themodel-Static, the internal structures of the HSTs remain fixed throughout the data stream, meaning that the counters inside the leaf nodes are not updated after initialization. In contrast, \themodel-Dynamic allows for continuous updates to the HSTs, enabling them to adapt to changes in the data stream.

As noted in~\cite{DBLP:conf/ijcai/TanTL11}, HSTs perform well when anomalies in the data stream are relatively sparse. However, their effectiveness diminishes when a large number of anomalies occur, as this can distort the counts within the tree ensemble. \themodel-Static mitigates this issue by preventing continuous updates and instead relying on a fixed subsample of the data. Within the algorithm, line 13 specifies where an update may occur when the \themodel-Dynamic variant is employed.

In the algorithm, we also incorporate a \emph{cache} memory of predefined size  $C$  to expedite the computation of anomaly scores for edges $e_t = (s,d,t)$ that have previously appeared. Specifically, if there exists an earlier occurrence $e_{t'} = (s,d,t') \in \mathcal{G}$ with $t' < t$, the cache prevents redundant recalculations. This optimization is particularly beneficial in scenarios such as burst traffic or denial-of-service (DoS) attacks, where a peer repeatedly attempts to connect to the same device within a short timeframe. To manage stored entries efficiently, the cache can implement various update policies, including FIFO (First-In, First-Out), LRU (Least Recently Used), or full replacement.


\begin{algorithm}
\caption{Streaming pipeline}\label{alg:streaming}
\begin{algorithmic}[1]
\State \textbf{Input:} pre-trained $\GNN^{enc}$, $\HST^e$, $\HST^v$, $\mathcal{G}$, cache size $C$
\State \textbf{Output:} $Scores$
\State $\Cache \gets \{\}$
\State $Scores \gets []$
\While{new edge $e_t=(s, d, t)$ occurs in $\mathcal{G}$}
    \State $G_{t}$ $\gets$ Update graph $G_{t-1}$
    \If {$(s,d) \in \Cache$}
        \State $\score^{e_t} \gets \Cache[(s,d)]$
    \Else
        \State $h^{s}$, $h^{d}$ $\gets \GNN^{enc}(G_t)$
        \State $h^{e_t} \gets \phi(h^{s}, h^{d})$ \Comment{Equations~\ref{eq:edgemean}/\ref{eq:edgesub}}
        \State $\score^{e_t} \gets score(e_t)$ \Comment{Equation~\ref{eq:anomalyscore}}
        \State Update $HST^e$, $HST^v$
        \State $\Cache[(s,d)] \gets score^{e_t}$
    \EndIf
\If {$|\Cache| > C$}
    \State Update $Cache$
\EndIf

    \State $Scores$.insert($score^{e_t}$)
\EndWhile
\end{algorithmic}
\end{algorithm}

\subsection{Validation-based Adaptive thresholding}\label{subsec:threshold}
Algorithm~\ref{alg:streaming} enables the computation of the anomaly score for each edge in the stream, within a range from $0$ (unlikely) to $1$ (very likely). For the choice of the appropriate threshold $\tau$, we devise a validation-based strategy that relies on a validation set $\{(e_{t_1}, l_1), \ldots, (e_{t_n},l_n)\}$, where anomaly labels are known for each edge. For each of these edges we can compute the anomaly score by selecting the associated temporal graph and computing the scores provided by the GNN embeddings in the HSTs computed by algorithm~\ref{alg:streaming}. The result is hence a set of pairs $D= \{(\score^{e_{t_1}}, l_1), ..., (\score^{e_{t_n}}, l_n)\}$.
The optimal threshold can be obtained by considering the partition that provides the minimal variance in the labels obtained by the two resulting subsets. For this, we exploit the Gini Diversity Index~\cite{BreiFrieStonOlsh84}. 
Let $D_1, D_2$ be two partitions of $D$, where $D_1 = \{(s, l)\in D | s \leq \tau \}$ and $D^2 = D - D_1$.
We then compute the optimal threshold $\tau^*$ by minimizing the formula
\begin{equation}
    \tau^* = \argmin_{\tau}\ \frac{|D_1|}{|D|} \times \mathrm{Gini}(D_1) + \frac{|D_2|}{|D|} \times \mathrm{Gini}(D_2), 
    \label{eq:argmingini}
\end{equation}
where 
$\mathrm{Gini}(S) = 1 - (p_{+}^2 + p_{-}^2)$ and $p_{+}$ (resp. $p_{-}$) is the fraction of positive (resp. negative) labels in $S$. 





\section{Analysis}
\label{sec:theoretical}





\themodel  relies on theoretical and empirical guarantees provided by its two main components, namely GNN and HST. First, as explained in Section~\ref{sec:approach}, $\GNN^{enc}$ exploits a GraphSAGE model. The latter has been shown to provide meaningful representations of nodes and edges that capture the underlying graph structure~\cite{hamilton2018inductive}. This is also true when the feature matrix $X_t$ is naively represented as a set of unique identifiers~\cite[Corollary 2]{hamilton2018inductive}. As a consequence, equipping \themodel with this architecture ensures that $\GNN^{enc}$ learns local graph structures and generates meaningful node and edge embeddings.


A second aspects is concerned with the ability of Half-Space Trees models to produce proper anomaly scores. As already mentioned above,  HST assumes that anomalies are rare within the dataset, as empirically shown in~\cite{DBLP:conf/ijcai/TanTL11}.
Consequently, we expect \themodel to work well in situations where the graph streams exhibits a limited amount of anomalies. Surprisingly, the framework produces  high-quality results even when this property does not occur in the data, as we will show later in Section~\ref{sec:experiments}.

We next analyze the computational complexity of the proposed approach. For this, we observe that the two main components of the primary loop in Algorithm~\ref{alg:streaming} (GNN encoding and HST exploitation and update) are executed sequentially. Let $G_t$ contain $n$ nodes, and assume GraphSAGE is configured with $\kappa$ layers, a recursive neighborhood sampling of $r$ neighbors, and a batch size of $s$. 
According to~\cite{9046288}, the time and space complexity of the GraphSAGE model are $O(r^{\kappa}nK^2)$ and $O(sr^{\kappa}K + \kappa K^2)$, respectively. In the space complexity expression, the term $sr^{\kappa}K$ accounts for storing the embeddings of sampled neighbor nodes across all layers, while $\kappa K^2$ represents the storage required for the model parameters.

Notably, within algorithm~\ref{alg:streaming}, GraphSAGE is used to embed only two individual nodes. The final embedding of a single node only needs to consider its $r$ neighbors. Since each layer aggregates and concatenates information from neighboring nodes, these dependencies propagate backward through multiple layers. Consequently, with an optimized implementation, the time complexity for computing the embedding of a single node is $O(r^{\kappa}K^2)$, which directly corresponds to the computational cost of step 10 in the algorithm.


The complexity of the HST algorithm primarily depends on the maximum tree height $h$, the number of trees $\beta$, and the window size $\psi$. As discussed in~\cite{DBLP:conf/ijcai/TanTL11}, the time complexity is $O(\beta(h + \psi))$ and the space complexity is $O(\beta 2^h)$.

By combining these results, the overall complexity for each iteration of Algorithm~\ref{alg:streaming} is $O(r^{\kappa}K^2 + \beta(h + \psi))$ in time and $O(sr^{\kappa}K + \kappa K^2 + \beta 2^h)$ in space. Assuming that $h$, $\psi$, $\beta$, $\kappa$, $s$, $r$, and $K$ are bounded, the overall complexity is determined solely by the number of nodes in the current graph $G_t$.

In practice, by assuming $r$ and $k$ bounded,  the complexity of \themodel\ is constant for each edge, offering a crucial feature in the edge streaming context. This property ensures that the model can efficiently handle large and continuously growing graphs without an exponential increase in computational cost. This theoretical advantage is further empirically validated through experimental results, as demonstrated in Section~\ref{sec:experiments}.


\section{Experiments}
\label{sec:experiments}
We evaluate the capabilities of the proposed framework through a series of experiments aimed at answering the following research questions:
\begin{description}
    \item[RQ1:] Can \themodel\ be used to recognize anomalies in edge streams? How does it compare to SOTA models?
    \item[RQ2:] How effective is the adaptive thresholding method in discriminating between normal and anomalous edges?
    \item[RQ3:] How does \themodel behave when facing different categories of anomalies?
    \item[RQ4:] How do the components of \themodel affect its capability to detect anomalies? How robust is the framework?
\end{description}

We discuss the results for both variants of the proposed approach, namely \themodel-Static and \themodel-Dynamic. Our implementation is developed using both PyTorch Geometric~\cite{Fey/Lenssen/2019} and River~\cite{montiel2021river} frameworks. All experiments were performed on an NVIDIA DGX equipped with 4 V100 GPUs.
The code used to perform the experiments is publicly available\footnote{\url{https://anonymous.4open.science/r/ARES-4573}}.

\subsection{Experimental Setting}
\paragraph{\textbf{Competitors}.}
We compare our results with three state-of-the-art models, discussed in the following. 
\begin{itemize}[leftmargin=*]
    \item MIDAS~\cite{DBLP:journals/tkdd/BhatiaLHYSF22} exploits count-min sketches~\cite{DBLP:journals/jal/CormodeM05} and a chi-squared test to evaluate the degree of abnormality for each edge. We consider two further variants of MIDAS: MIDAS-R~\cite{DBLP:conf/aaai/0001HYSF20} that incorporates temporal and spatial relations, and MIDAS-F, which improves MIDAS-R by filtering anomalies to prevent negatively altering the internal structures.
    \item AnoGraph~\cite{DBLP:conf/kdd/0001WKSYH23} expands the concept of count-min sketches to strengthen the ability of representing multidimensional inputs. 
    In our experiments, we used the two variants focusing on edge anomalies: AnoEdge-G and AnoEdge-L. 
    \item SLADE-H~\cite{DBLP:conf/kdd/LeeKS24} employs a Temporal Graph Neural Network architecture for detecting anomalous edges. The model assigns anomaly scores by measuring the cosine similarity between the node’s current and previous memory vectors, which are used to track the nodes' evolution.
\end{itemize}
To ensure a fair comparison, we used the source code provided by the authors and performed a tuning of the hyperparameters according to the findings described in the original papers. 

\paragraph{\textbf{Metrics}.}
We report two widely used metrics in anomaly detection: the Area Under the ROC Curve (ROC-AUC) and Average Precision (AP). The ROC-AUC measures the model's ability to distinguish between classes by computing the true positive rate (TPR) against the false positive rate (FPR) across different threshold values. Similarly, the AP summarizes the precision-recall curve by calculating precision and recall at various thresholds, providing insight into the model's performance in imbalanced scenarios.

In addition, to answer \textbf{RQ2} and evaluate the model capabilities under the thresholding mechanism, we report the F1-Score and Balanced Accuracy (B-Accuracy)  metrics. Both metrics are calculated using the threshold determined by the adaptive thresholding methods described in section~\ref{subsec:threshold}. For the competitors, we evaluate multiple threshold values and report the results for the threshold that yields the highest F1-Score on the validation set.



\paragraph{\textbf{Datasets}.}
Our evaluation is based on the following datasets:
\begin{itemize}[leftmargin=*]
    \item DARPA~\cite{DARPA}, one of the most widely used benchmarks, featuring various categories of cyber threats, including Denial-of-Service, Remote-to-Local, User-to-Root, and Surveillance attacks.
    \item ISCX2012\cite{ISCX} and CIC-IDS2017\cite{IDS2018}, provided by the University of New Brunswick (UNB), Canada. ISCX2012 includes network traffic for protocols such as HTTP, SMTP, SSH, IMAP, POP3, and FTP, while CIC-IDS2017 covers a range of attack types, including DoS, DDoS, Brute Force, XSS, SQL Injection, Infiltration, Port Scan, and Botnet.
    \item UNSW-NB15~\cite{UNSW-NB15}, which consists of network traffic including both real and synthetic attack activities.
    \item CTU-13~\cite{CTU-13}, a botnet traffic dataset containing 13 distinct attack scenarios. We excluded scenarios with fewer than 1 million edges or that closely resembled other attack types. The final selected scenarios are 1, 10, and 13.
\end{itemize} 

Table~\ref{tab:appendix_datasets} (Appendix) summarizes the statistics of the seven real-world datasets. Each dataset is temporally split: $60\%$ for training, $20\%$ for validation, and $20\%$ for testing. The training set is used to train the GAE and initialize $\HST^e$ and $\HST^v$, and similarly for training SLADE-H and initializing MIDAS/AnoGraph structures.

We use the anomaly scores on the validation set for tuning the hyper-parameters. In \themodel, the edge representation (Eq.\ref{eq:edgemean} vs. Eq.\ref{eq:edgesub}) is also selected via validation. Final results are computed on the test set using Algorithm~\ref{alg:streaming}.


\subsection{Results}
\paragraph{\textbf{Recognizing anomalies}}
To answer \textbf{RQ1} and compare the results of \themodel against the competitors, we repeat the experiments using nine different seeds. The resulting metrics are compared using the Wilcoxon test with a confidence level of $95\%$. 
Table \ref{tab:auc_ap_results} report the comparisons. Bold and underlined values denote best and second-best, respectively. Statistical ties are represented by multiple bold/underlined values. 

\themodel\ demonstrates prominent results in terms of ROC-AUC and AP, exhibiting its capabilities in separating anomalies from normal edges. The only exception is on CIC-IDS2017 where MIDAS-F shows better scores both in terms of AUC and AP. The row relative to "Performance Gain"  quantifies the relative improvement of \themodel by displaying the percentage difference between its best results and those achieved by competing methods, offering a direct comparison of performance improvement.
In general, \themodel consistently outperforms all the competitors, achieving improvements up to $+12.8\%$ on Scenario 1 and $+23.5\%$ on Scenario 10 in terms of ROC-AUC and AP, respectively.

\begin{figure}
    \begin{center}
    \includegraphics[width=.5\textwidth]{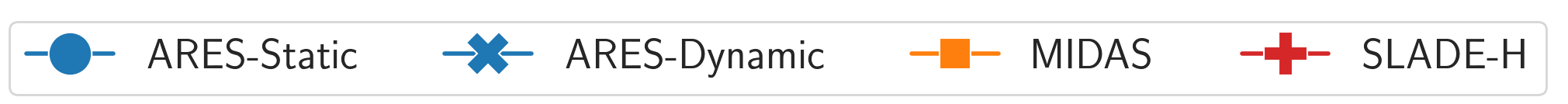}\\
    \includegraphics[width=.32\textwidth]{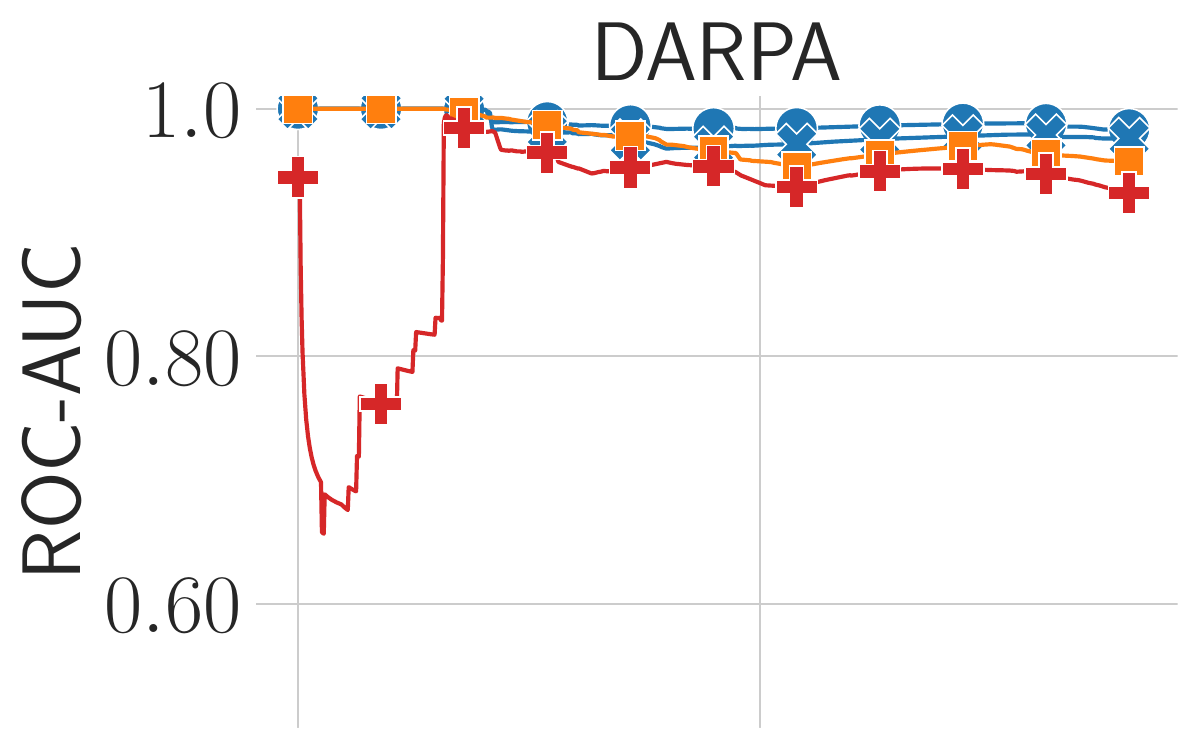}
    \includegraphics[width=.32\textwidth]{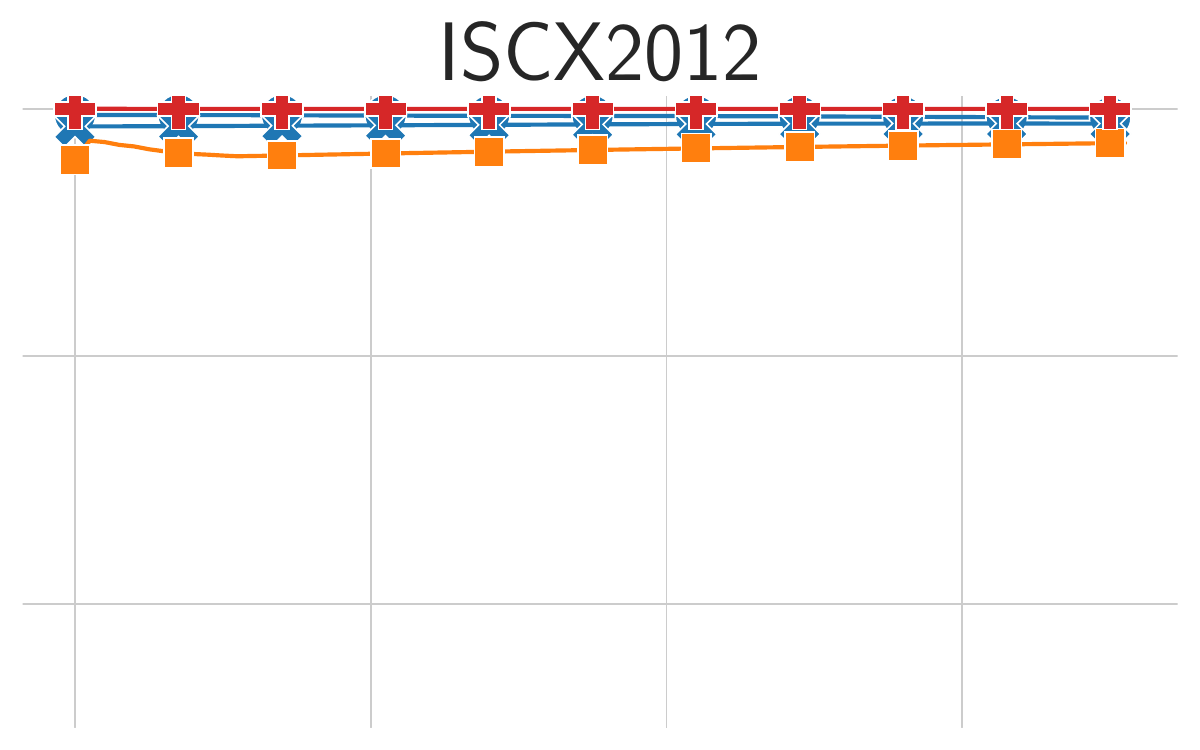}
    \includegraphics[width=.32\textwidth]{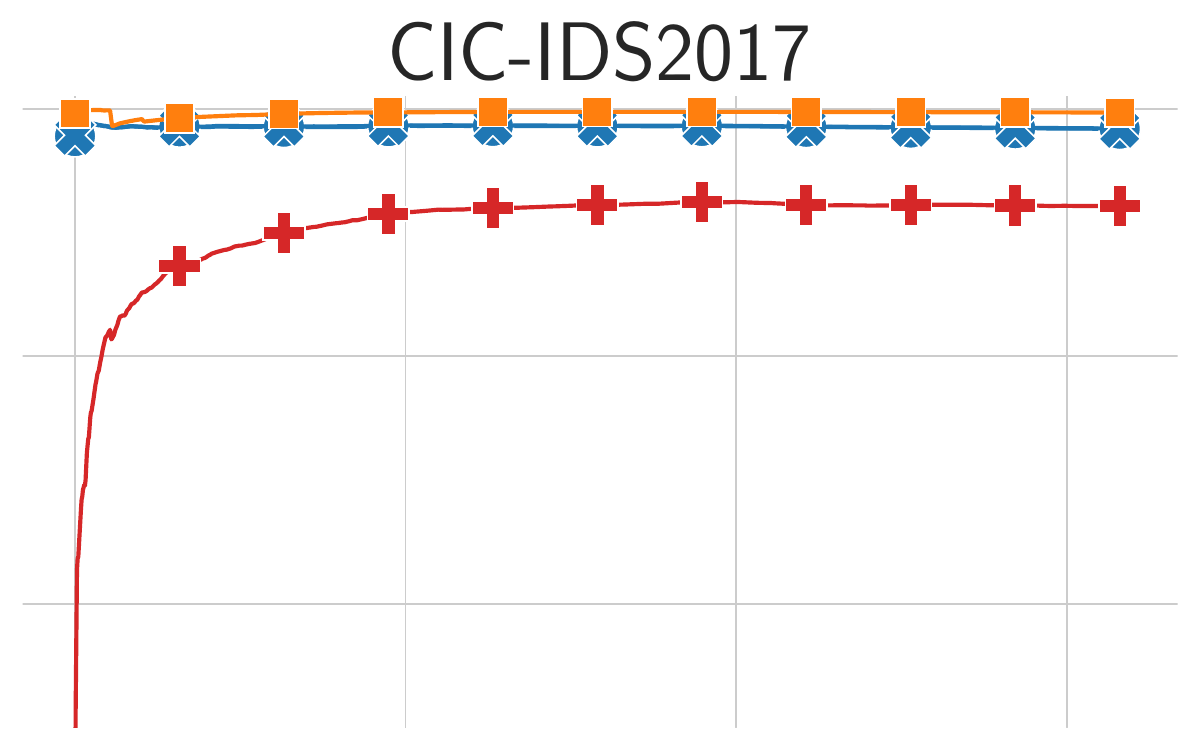}
    \includegraphics[width=.32\textwidth]{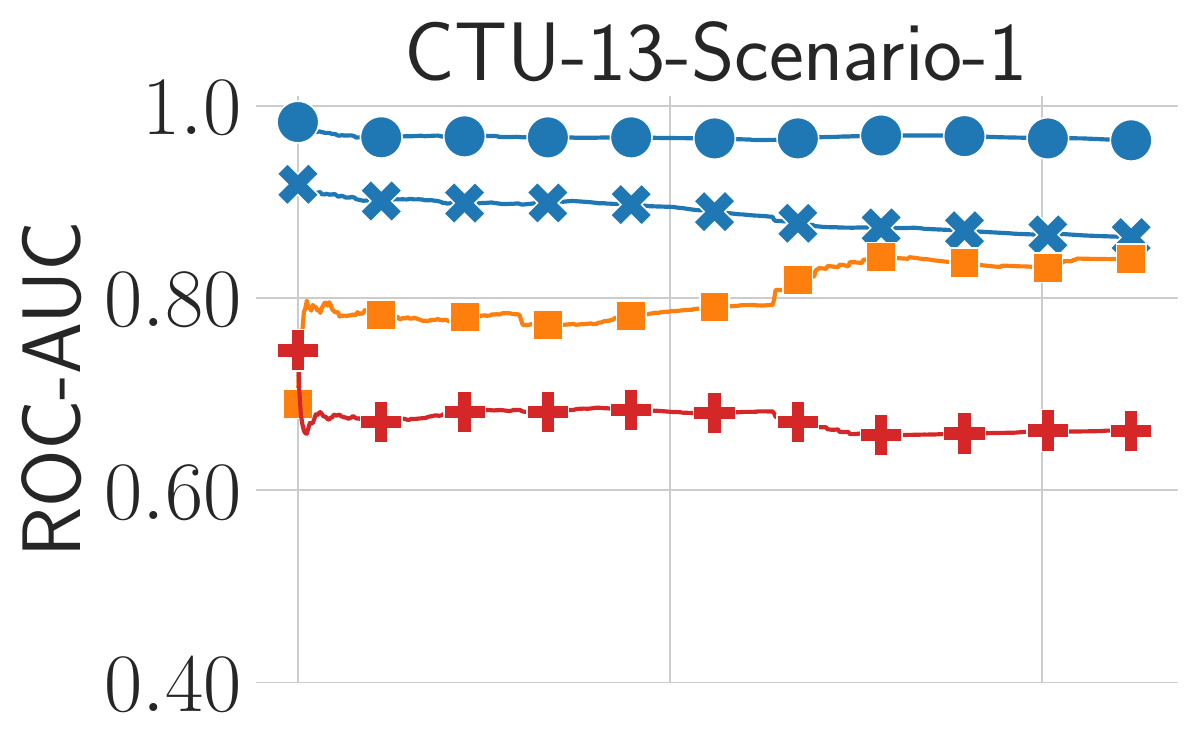}
    \includegraphics[width=.32\textwidth]{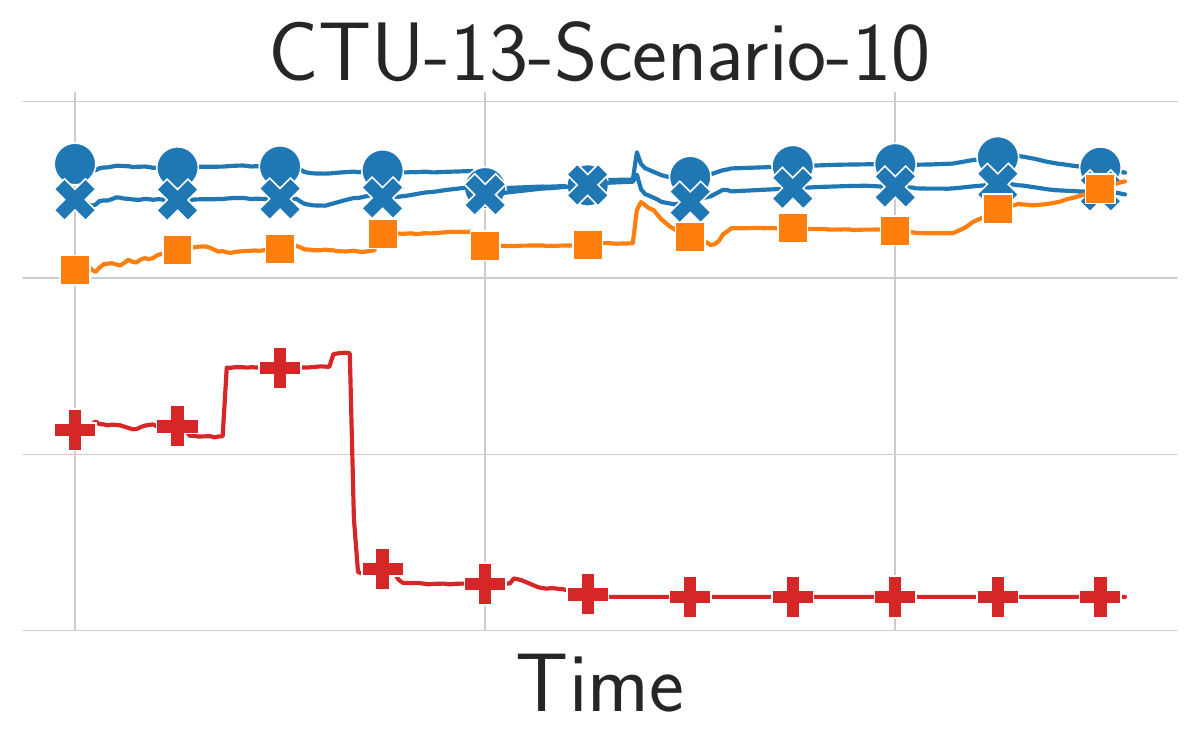}
    \includegraphics[width=.32\textwidth]{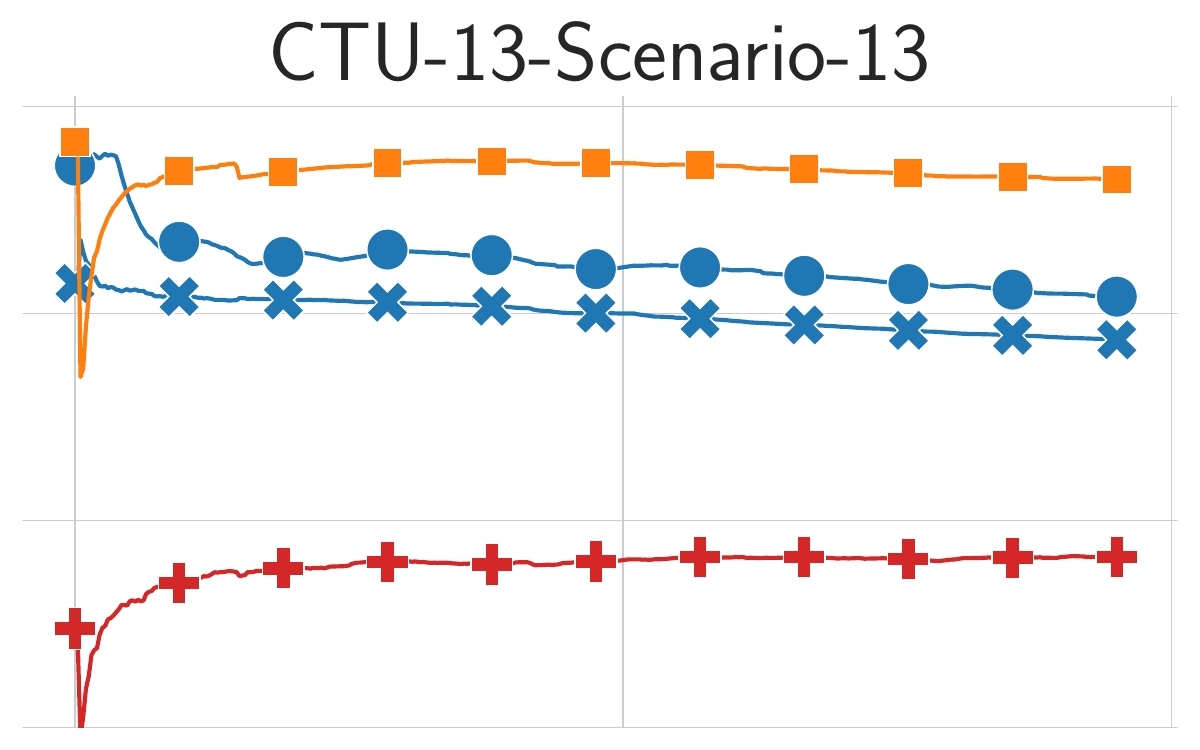}
    \includegraphics[width=.33\textwidth]{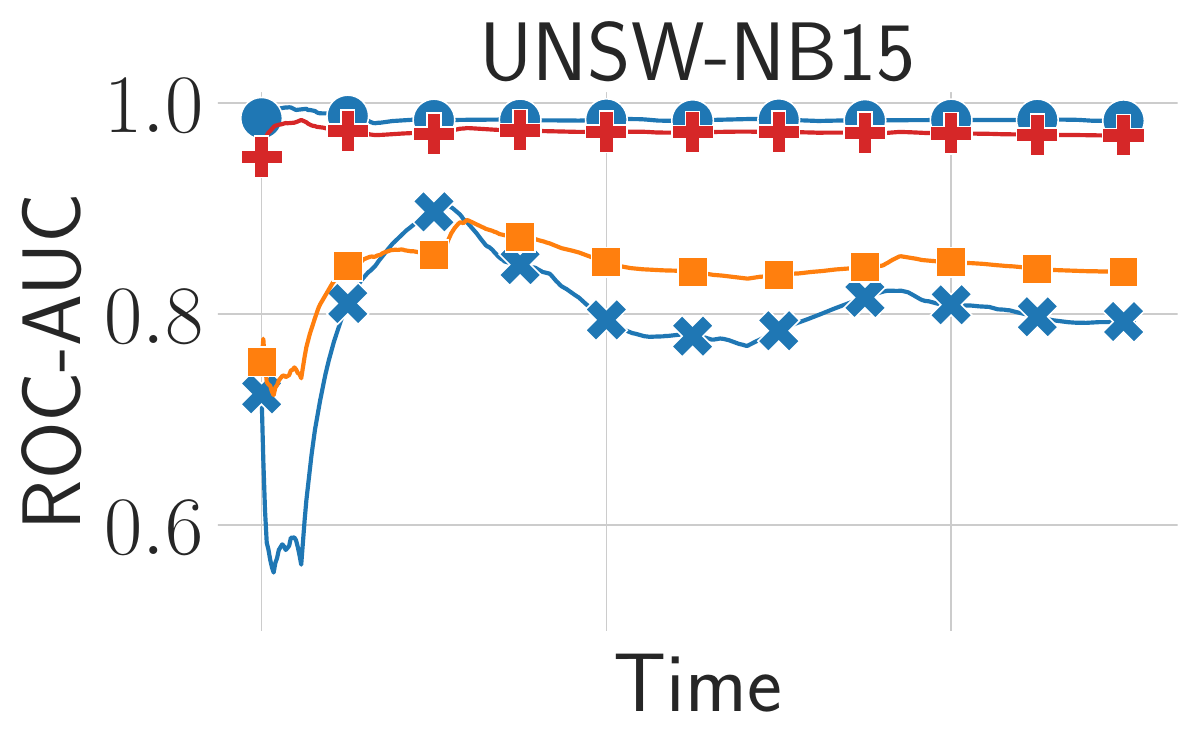}
    \caption{Comparisons of ROC-AUC scores over time among \themodel-Static, \themodel-Dynamic, MIDAS, and SLADE-H. 
    The best variant of MIDAS is take into account for each dataset, as reported in Table~\ref{tab:auc_ap_results}}
    \label{fig:roc_over_time}
    \end{center}
\end{figure}

\paragraph{\textbf{Robustness to concept drifts.}}
To further validate \themodel, we assess its performance over time using the AUC-ROC metric, as illustrated in Figure~\ref{fig:roc_over_time}. This evaluation highlights the model’s robustness in the presence of concept drift—an inherent challenge in streaming scenarios. The results demonstrate that \themodel consistently maintains stable performance, outperforming both MIDAS and SLADE-H. We omit results for AnoEdge, as its performance is significantly lower in comparison.

The robustness of \themodel stems from the synergy of \themodel’s core components. As previously discussed, the graph $G_t$ is continuously updated, enabling the GNN to generate node and edge embeddings that adapt to the evolving data stream. Simultaneously, HST incrementally updates its internal structures with incoming edges, ensuring that anomaly scores are reflective of the most recent data.
\begin{table*}[ht!]
\centering
\resizebox{\textwidth}{!}{
\begin{tabular}{@{}ccccccccccccccc@{}}
\toprule
\multirow{3}{*}{Model} & \multicolumn{2}{c}{\multirow{2}{*}{DARPA}} & \multicolumn{2}{c}{\multirow{2}{*}{UNSW-NB15}} & \multicolumn{2}{c}{\multirow{2}{*}{ISCX2012}} & \multicolumn{2}{c}{\multirow{2}{*}{CIC-IDS2017}} & \multicolumn{6}{c}{CTU-13} \\ \cmidrule(l){10-15} 
 & \multicolumn{2}{c}{} & \multicolumn{2}{c}{} & \multicolumn{2}{c}{} & \multicolumn{2}{c}{} & \multicolumn{2}{c}{Scenario 1} & \multicolumn{2}{c}{Scenario 10} & \multicolumn{2}{c}{Scenario 13} \\ \cmidrule(l){2-15} 
 & ROC-AUC & \multicolumn{1}{c|}{AP} & ROC-AUC & \multicolumn{1}{c|}{AP} & ROC-AUC & \multicolumn{1}{c|}{AP} & ROC-AUC & \multicolumn{1}{c|}{AP} & ROC-AUC & \multicolumn{1}{c|}{AP} & ROC-AUC & \multicolumn{1}{c|}{AP} & ROC-AUC & AP \\ \midrule
MIDAS-F & $\underline{0.957 \pm 0.004}$ & $\underline{0.977 \pm 0.002}$ & $0.793 \pm 0.001$ & $0.337 \pm 0.001$ & \underline{$0.972 \pm 0.002$} & $0.275 \pm 0.015$ & \bm{$0.997 \pm 0.000$} & \bm{$0.968 \pm 0.012$} & \underline{$0.841 \pm 0.041$} & $0.077 \pm 0.015$ & $0.911 \pm 0.022$ & \underline{$0.477 \pm 0.074$} & \bm{$0.929 \pm 0.051$} & \underline{$0.131 \pm 0.049$} \\
MIDAS-R & $0.840 \pm 0.003$ & $0.921 \pm 0.001$ & $0.840 \pm 0.002$ & $0.469 \pm 0.001$ & $0.771 \pm 0.021$ & $0.067 \pm 0.006$ & $0.791 \pm 0.006$ & $0.130 \pm 0.004$ & $0.618 \pm 0.061$ & $0.031 \pm 0.007$ & $0.294 \pm 0.000$ & $0.085 \pm 0.000$ & $0.342 \pm 0.010$ & $0.012 \pm 0.001$ \\
MIDAS & $0.781 \pm 0.006$ & $0.889 \pm 0.003$ & $0.815 \pm 0.005$ & $0.444 \pm 0.002$ & $0.597 \pm 0.001$ & $0.033 \pm 0.001$ & $0.544 \pm 0.027$ & $0.071 \pm 0.005$ & $0.523 \pm 0.007$ & $0.017 \pm 0.001$ & $0.305 \pm 0.002$ & $0.087 \pm 0.002$ & $0.389 \pm 0.001$ & $0.013 \pm 0.000$ \\
AnoEdge-G & $0.895 \pm 0.009$ & $0.948 \pm 0.004$ & $0.756 \pm 0.012$ & $0.423 \pm 0.013$ & $0.867 \pm 0.014$ & $0.097 \pm 0.004$ & $0.971 \pm 0.010$ & $0.586 \pm 0.088$ & $0.436 \pm 0.006$ & $0.015 \pm 0.001$ & $0.661 \pm 0.000$ & $0.153 \pm 0.001$ & $0.294 \pm 0.011$ & $0.011 \pm 0.001$ \\
AnoEdge-L & $0.909 \pm 0.006$ & $0.953 \pm 0.003$ & $0.752 \pm 0.019$ & $0.437 \pm 0.031$ & $0.893 \pm 0.025$ & $0.113 \pm 0.026$ & $0.980 \pm 0.005$ & $0.627 \pm 0.066$ & $0.484 \pm 0.017$ & $0.041 \pm 0.003$ & $0.696 \pm 0.016$ & $0.180 \pm 0.017$ & $0.315 \pm 0.021$ & $0.013 \pm 0.001$ \\ \midrule
SLADE-H & $0.931 \pm 0.042$ & $0.958 \pm 0.031$ & $\underline{0.971 \pm 0.021}$ & \underline{$0.867 \pm 0.096$} & \bm{$0.999 \pm 0.001$} & \bm{$0.975 \pm 0.021$} & $0.921 \pm 0.096$ & $0.413 \pm 0.201$ & $0.662 \pm 0.178$ & $0.035 \pm 0.019$ & $0.438 \pm 0.216$ & $0.114 \pm 0.041$ & $0.565 \pm 0.077$ & $0.019 \pm 0.005$ \\ \midrule
\themodel-Static & \bm{$0.985 \pm 0.008$} & \bm{$0.991 \pm 0.005$} & \bm{$0.985 \pm 0.002$} & \bm{$0.892 \pm 0.014$} & \bm{$0.999 \pm 0.000$} & $0.942 \pm 0.007$ & \underline{$0.984 \pm 0.001$} & \underline{$0.806 \pm 0.025$} & \bm{$0.969 \pm 0.004$} & \bm{$0.291 \pm 0.014$} & \bm{$0.956 \pm 0.019$} & \bm{$0.712 \pm 0.086$} & \bm{$0.948 \pm 0.049$} & \bm{$0.249 \pm 0.091$} \\
\themodel-Dynamic & \bm{$0.982 \pm 0.005$} & \bm{$0.990 \pm 0.002$} & $0.876 \pm 0.002$ & $0.486 \pm 0.008$ & \bm{$0.999 \pm 0.000$} & \underline{$0.961 \pm 0.017$} & \underline{$0.984 \pm 0.001$} & \underline{$0.816 \pm 0.009$} & \underline{$0.863 \pm 0.103$} & \underline{$0.090 \pm 0.049$} & \underline{$0.940 \pm 0.015$} & \bm{$0.631 \pm 0.064$} & \underline{$0.853 \pm 0.039$} & \underline{$0.094 \pm 0.041$} \\ \midrule
Performance Gain (\%) & $+2.8$ & $+1.4$ & $+1.4$ & $+2.5$ & $0.0$ & $-1.4$ & $-1.3$ & $-15.2$ & $+12.8$ & $+21.4$ & $+4.5$ & $+23.5$ & $+1.9$ & $+11.8$ \\\bottomrule
\end{tabular}
}
\caption{Evaluation of \themodel’s performance against competitors in identifying anomalies across different datasets.}
\label{tab:auc_ap_results}
\end{table*}

\begin{table*}[!ht]
\centering
\resizebox{\textwidth}{!}{
\begin{tabular}{@{}ccccccccccccccc@{}}
\toprule
\multirow{3}{*}{Model} & \multicolumn{2}{c}{\multirow{2}{*}{DARPA}} & \multicolumn{2}{c}{\multirow{2}{*}{UNSW-NB15}} & \multicolumn{2}{c}{\multirow{2}{*}{ISCX2012}} & \multicolumn{2}{c}{\multirow{2}{*}{CIC-IDS2017}} & \multicolumn{6}{c}{CTU-13} \\ \cmidrule(l){10-15} 
 & \multicolumn{2}{c}{} & \multicolumn{2}{c}{} & \multicolumn{2}{c}{} & \multicolumn{2}{c}{} & \multicolumn{2}{c}{Scenario 1} & \multicolumn{2}{c}{Scenario 10} & \multicolumn{2}{c}{Scenario 13} \\ \cmidrule(l){2-15} 
 & F1-Score & \multicolumn{1}{c|}{B-Accuracy} & F1-Score & \multicolumn{1}{c|}{B-Accuracy} & F1-Score & \multicolumn{1}{c|}{B-Accuracy} & F1-Score & \multicolumn{1}{c|}{B-Accuracy} & F1-Score & \multicolumn{1}{c|}{B-Accuracy} & F1-Score & \multicolumn{1}{c|}{B-Accuracy} & F1-Score & B-Accuracy \\ \midrule
MIDAS-F & $0.804 \pm 0.003$ & $0.595 \pm 0.008$ & $0.636 \pm 0.004$ & \underline{$0.834 \pm 0.005$} & $0.485 \pm 0.030$ & \underline{$0.974 \pm 0.003$} & \underline{$0.797 \pm 0.019$} & \underline{$0.980 \pm 0.002$} & $0.051 \pm 0.003$ & $0.630 \pm 0.040$ & $0.291 \pm 0.010$ & $0.667 \pm 0.015$ & $0.045 \pm 0.009$ & $0.639 \pm 0.090$ \\
MIDAS-R & $0.815 \pm 0.002$ & \underline{$0.835 \pm 0.004$} & $0.599 \pm 0.000$ & $0.799 \pm 0.001$ & $0.116 \pm 0.058$ & $0.573 \pm 0.048$ & $0.309 \pm 0.004$ & $0.832 \pm 0.003$ & $0.047 \pm 0.000$ & $0.619 \pm 0.000$ & $0.061 \pm 0.113$ & $0.285 \pm 0.187$ & $0.033 \pm 0.000$ & $0.518 \pm 0.000$ \\
MIDAS & $0.741 \pm 0.006$ & $0.754 \pm 0.012$ & $0.547 \pm 0.000$ & $0.737 \pm 0.000$ & $0.054 \pm 0.000$ & $0.553 \pm 0.004$ & $0.174 \pm 0.005$ & $0.619 \pm 0.005$ & $0.047 \pm 0.000$ & $0.622 \pm 0.000$ & $0.268 \pm 0.003$ & $0.625 \pm 0.007$ & $0.037 \pm 0.000$ & $0.583 \pm 0.000$ \\
AnoEdge-G & $0.809 \pm 0.000$ & $0.831 \pm 0.002$ & $0.530 \pm 0.032$ & $0.739 \pm 0.021$ & $0.056 \pm 0.002$ & $0.587 \pm 0.016$ & $0.644 \pm 0.047$ & $0.943 \pm 0.016$ & $0.034 \pm 0.001$ & $0.491 \pm 0.005$ & $0.220 \pm 0.002$ & $0.515 \pm 0.007$ & $0.032 \pm 0.000$ & $0.502 \pm 0.001$ \\
AnoEdge-L & $0.805 \pm 0.000$ & \underline{$0.835 \pm 0.001$} & $0.573 \pm 0.022$ & $0.773 \pm 0.015$ & $0.087 \pm 0.009$ & $0.738 \pm 0.035$ & $0.734 \pm 0.040$ & $0.942 \pm 0.020$ & $0.040 \pm 0.003$ & $0.520 \pm 0.009$ & \underline{$0.331 \pm 0.012$} & \underline{$0.679 \pm 0.018$} & $0.006 \pm 0.001$ & $0.404 \pm 0.05$ \\ \midrule
SLADE-H & $0.844 \pm 0.051$ & $0.702 \pm 0.146$ & \underline{$0.827 \pm 0.112$} & \underline{$0.897 \pm 0.076$} & \underline{$0.556 \pm 0.492$} & $0.778 \pm 0.247$ & $0.587 \pm 0.221$ & $0.881 \pm 0.162$ & $\underline{0.096 \pm 0.037}$ & \underline{$0.769 \pm 0.147$} & $0.095 \pm 0.172$ & $0.485 \pm 0.174$ & $0.060 \pm 0.009$ & \underline{$0.739 \pm 0.058$} \\ \midrule
\themodel-Static & \bm{$0.970 \pm 0.005$} & \bm{$0.968 \pm 0.001$} & \bm{$0.939 \pm 0.002$} & \bm{$0.984 \pm 0.001$} & \bm{$0.964 \pm 0.012$} & \bm{$0.999 \pm 0.000$} & \bm{$0.894 \pm 0.005$} & \bm{$0.991 \pm 0.000$} & \bm{$0.334 \pm 0.089$} & \bm{$0.957 \pm 0.022$} & \bm{$0.711 \pm 0.142$} & \bm{$0.849 \pm 0.095$} & \bm{$0.343 \pm 0.194$} & \bm{$0.867 \pm 0.203$} \\
\themodel-Dynamic & \underline{$0.962 \pm 0.006$} & \bm{$0.962 \pm 0.007$} & $0.622 \pm 0.055$ & $0.815 \pm 0.067$ & $0.437 \pm 0.482$ & $0.722 \pm 0.247$ & \bm{$0.899 \pm 0.005$} & \bm{$0.999 \pm 0.000$} & \underline{$0.104 \pm 0.109$} & $0.579 \pm 0.121$ & \bm{$0.612 \pm 0.092$} & \bm{$0.838 \pm 0.061$} & \underline{$0.127 \pm 0.110$} & $0.585 \pm 0.084$ \\ \midrule
Performance Gain (\%) & $+12.6$ & $+13.3$ & $+11.2$ & $+8.7$ & $+40.8$ & $+2.5$ & $+10.2$ & $+1.9$ & $+23.8$ & $+18.8$ & $+4.5$ & $+38.0$ & $+28.3$ & $+12.8$ \\\bottomrule
\end{tabular}
}
\caption{Evaluation of \themodel with thresholding discrimination. \themodel with adaptive thresholding is compared against the optimal threshold devised for each competitor.}
\label{tab:threshold_results}
\end{table*}

\paragraph{\textbf{Discerning between normal and abnormal}} The behavior of the model under the the adaptive thresholding method (and hence the answer to \textbf{RQ2}) can be observed through the values of F1, B-Accuracy and Precision/Recall reported in Table~\ref{tab:threshold_results}.
We observe robust results on both the static and dynamic versions of the model. \themodel-Dynamic exhibits minor capabilities in terms of recall, but overall both methods are stable in detecting anomalies. Notably, the proposed method provides a simple yet effective alternative to baseline approaches, which lack an algorithmic mechanism for computing thresholds.

\paragraph{\textbf{Detecting different cyber-attacks.}}
To investigate RQ3, we evaluate the ability of \themodel in detecting anomalies relative to specific attack scenarios. First, notice that the CTU-13 dataset variants already contain single different scenarios, referring to Click-fraud, UDP DDoS, and Port-scan attacks respectively. And in fact, as shown in Table~\ref{tab:auc_ap_results}, \themodel outperforms its competitors in detecting such attacks. 



We further conduct an in-depth analysis of accuracy scores for specific attack types in the DARPA and UNSW-NB15 (Table~\ref{tab:cyber_attacks_combined}) datasets, both of which encompass diverse anomaly patterns, including bursty and spike anomalies. In both cases, \themodel exhibits an  accuracy remains consistent across all available attack types.

\begin{table}[!ht]
\centering
\resizebox{0.5\columnwidth}{!}{
\begin{tabular}{@{}lccc@{}}
\toprule
Dataset & Cyber-Attack & ROC-AUC & AP \\ \midrule
\multirow{4}{*}{DARPA} 
& Denial-of-Service       & $0.989 \pm 0.008$    & $0.993 \pm 0.004$  \\
& Surveillance/Probing    & $0.998 \pm 0.001$    & $0.999 \pm 0.001$  \\
& Remote-to-Local         & $0.972 \pm 0.012$    & $0.982 \pm 0.008$  \\
& User-to-Root            & $1.000 \pm 0.000$    & $1.000 \pm 0.000$  \\ \midrule
\multirow{7}{*}{UNSW-NB15} 
& Exploits                & $0.984 \pm 0.001$    & $0.828 \pm 0.012$  \\
& Reconnaissance          & $0.984 \pm 0.001$    & $0.851 \pm 0.014$  \\
& DoS                     & $0.987 \pm 0.002$    & $0.786 \pm 0.022$  \\
& Shellcode               & $0.980 \pm 0.002$    & $0.840 \pm 0.017$  \\
& Fuzzers                 & $0.984 \pm 0.002$    & $0.870 \pm 0.014$  \\
& Worms                   & $0.991 \pm 0.004$    & $0.942 \pm 0.030$  \\
& Backdoors               & $0.980 \pm 0.003$    & $0.705 \pm 0.024$  \\ \bottomrule
\end{tabular}
}
\caption{Anomaly detection performance (ROC-AUC and AP) for specific cyber-attack types in the DARPA and UNSW-NB15 datasets.}
\label{tab:cyber_attacks_combined}
\end{table}

\begin{table*}[!ht]
\centering
\resizebox{\textwidth}{!}{
\begin{tabular}{@{}cccccccc@{}}
\toprule
Model & DARPA & UNSW-NB15 & ISCX2012 & CIC-IDS2017 & CTU-13 Scenario 1 & CTU-13 Scenario 10 & CTU-13 Scenario 13 \\ \midrule
MIDAS-F           & $0.39$  & $0.43$  & $1.26$  & $0.58$  & $17.02$ & $8.32$  & $12.24$  \\
MIDAS-R           & $0.21$  & $0.10$  & $0.21$  & $0.42$  & $3.69$  & $1.24$  & $1.89$   \\
MIDAS             & $0.03$  & $0.02$  & $0.02$  & $0.06$  & $0.24$  & $0.08$  & $0.11$   \\
AnoEdge-G         & $39.41$ & $23.17$ & $13.62$ & $96.55$ & $31.51$ & $11.87$ & $24.09$  \\
AnoEdge-L         & $0.33$  & $0.32$  & $0.22$  & $0.64$  & $2.29$  & $0.71$  & $0.93$   \\
\midrule
SLADE-H & $763.09$  & $189.63$  & $29.83$  & $1539.85$  & $505.57$  & $67.07$  & $230.36$   \\
\midrule
\themodel-Static  & $7.83$ & $5.40$ & $14.13$  & $51.36$ & $59.52$ & $44.85$ & $108.26$ \\
\themodel-Dynamic & $10.84$ & $6.11$ & $6.76$  & $90.76$ & $74.33$ & $51.08$ & $70.20$  \\ \bottomrule
\end{tabular}
}
\caption{Running times (in seconds) across the used datasets}
\label{tab:timings}
\end{table*}


\begin{figure}[!ht]
    \centering
    \includegraphics[width=0.4\textwidth]{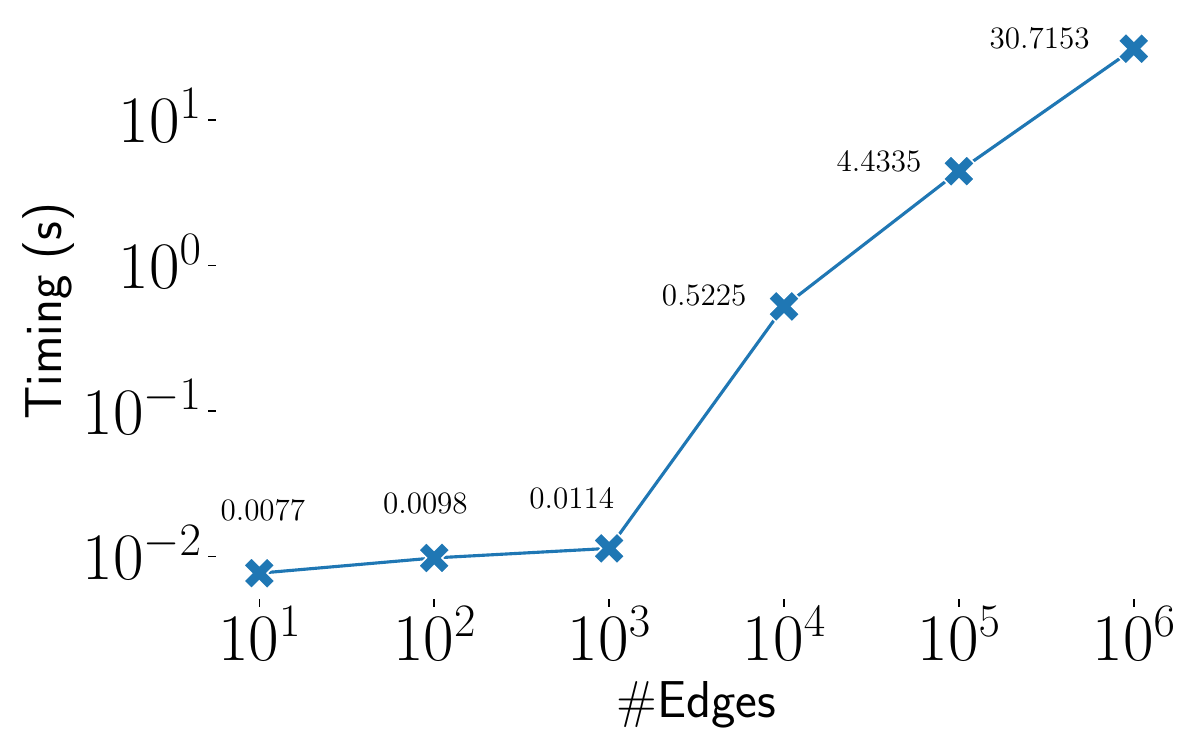}
    \caption{ARES-Static timings on CIC-IDS2017 dataset as the number of edges increases}
    \label{fig:scalability}
\end{figure}

\paragraph{\textbf{Timings}.} Table \ref{tab:timings} presents the running times for all datasets. Compared to rule-based models, \themodel has higher computational costs due to the integration of the GNN and HST architectures. However, unlike SLADE-H, the proposed model leverages a lightweight yet powerful GNN, leading to significant computational speed-ups. This efficiency gain makes \themodel a well-balanced solution, striking an optimal tradeoff between accuracy and performance.
The trade-off in computational cost is justified by the significant accuracy improvements ARES offers across diverse datasets. The model consistently outperforms MIDAS and AnoEdge in 6 out of 7 datasets in both AUC and AP metrics, highlighting its superior capability to detect subtle and complex anomalies in evolving graph structures. In applications where detection performance is critical, the additional computational overhead becomes a justified and worthwhile investment.
Figure \ref{fig:scalability} presents a scalability analysis for increasing edge stream sizes. The running times exhibit linear growth as the number of edges increases. Notably, an overhead emerges when the number of edges exceeds $10^3$, due to the \(\GNN^{enc}\) component’s need to process more batches and large adjacency matrices. Notwithstanding, the empirical result confirms the theoretical assessment discussed in Section~\ref{sec:theoretical}.

%


\paragraph{\textbf{Ablation study and Statistical robustness.}} To answer \textbf{RQ4}, we conduct an ablation study on the two core components of \themodel: Half-Space Trees (HST) and GraphSAGE embeddings.
First, we evaluate the contribution of HST by comparing it with RRCF~\cite{DBLP:conf/icml/GuhaMRS16} and a baseline GNN Encoder-Decoder (GNN-RecErr), which uses the reconstruction error as the anomaly score after previous work~\cite{DBLP:journals/tkde/MaWXYZSXA23}. As shown in Table~\ref{tab:darpa_ablation}, HST provides significantly better accuracy. Notably, it also outperforms RRCF in inference speed (see Table~\ref{tab:timings_hst_rrcf} in the Appendix).
Second, we assess the impact of the embedding model by testing different GNNs with HST. GraphSAGE consistently achieves the best performance, outperforming both GCN and GAT (Table~\ref{tab:darpa_ablation}). These results confirm the importance of high-quality graph embeddings and demonstrate the strong synergy between GraphSAGE and HST within \themodel. Due to space limitations, we only provide the results for the CTU-13 Scenario 1 dataset. More details on the ablation study are available in Appendix~\ref{sec:appendix_ablation}.

\begin{table}[!ht]
\centering
\resizebox{0.5\columnwidth}{!}{
\begin{tabular}{@{}ccc@{}}
\toprule
Model & ROC-AUC & AP \\ \midrule
GNN-RecErr & $0.394 \pm 0.000$ & $0.022 \pm 0.000$ \\ \midrule
GCN+HST & $0.707 \pm 0.173$ & $0.066 \pm 0.032$ \\
GAT+HST & $0.654 \pm 0.009$ & $0.034 \pm 0.005$ \\ \midrule
GraphSAGE+RRCF & $0.804 \pm 0.064$ & $0.072 \pm 0.025$ \\ \midrule
\themodel & \bm{$0.969 \pm 0.004$} & \bm{$0.291 \pm 0.014$} \\ \bottomrule
\end{tabular}
}
\caption{Ablation study on the CTU-13 Scenario 1 dataset demonstrates the effectiveness of combining GraphSAGE embeddings with Half-Space Trees.}
\label{tab:darpa_ablation}
\end{table}

We also measure the statistical robustness of our findings. Table~\ref{tab:subsampling_darpa} presents the ROC-AUC and AP performance of \themodel compared to competing models, on different subsamples of the DARPA dataset. The latter is particularly challenging as it comprises a large number ($59.9\%$) of edge anomalies. For each seed used, we uniformly sampled $50\%$ of the test set and computed the anomaly detection scores. As we can see, the results are consistent and stable along all runs, witnessing the robustness of the proposed model in handling such a challenging dataset. Experiments concerning the other datasets are in Appendix~\ref{sec:appendix_ablation}.

\begin{table}[!ht]
\centering
\resizebox{0.5\columnwidth}{!}{
\begin{tabular}{@{}ccc@{}}
\toprule
Model & ROC-AUC & AP\\ \midrule
MIDAS-F & $0.961 \pm 0.005$ & $0.979 \pm 0.003$ \\
AnoEdge-L & $0.910 \pm 0.007$ & $0.955 \pm 0.003$  \\ 
SLADE-H & $0.900 \pm 0.094$ & \bm{$0.912 \pm 0.091$} \\ \midrule
\themodel-Static & \bm{$0.991 \pm 0.005$} & \bm{$0.994 \pm 0.003$} \\
\themodel-Dynamic & \bm{$0.989 \pm 0.007$} & \bm{$0.994 \pm 0.003$} \\ \bottomrule
\end{tabular}
}
\caption{Comparison of \themodel against competitors by subsampling the dataset DARPA}
\label{tab:subsampling_darpa}
\end{table}


\section{Conclusion}
\label{sec:conclusions}

We proposed \themodel, an unsupervised anomaly detection method that combines Graph Neural Networks (GNNs) with Half-Space Trees to analyze edge streams efficiently, ensuring constant-time processing for each incoming edge. 
The model is equipped with a simple yet effective supervised algorithmic mechanism for determining a threshold to differentiate between normal and anomalous edges. 
Extensive empirical evaluation demonstrates its competitive advantage over state-of-the-art models.

There are several potential directions for improving this framework, which should be explored in future research. One key area is the development of more complex continual learning procedure to better update the GNN, which would enhance the quality of the embeddings over time. Additionally, methods to reduce computational overhead by accelerating GNN calculations should be considered. Another area for improvement is in the proposed thresholding method. Currently, \themodel uses a supervised approach, but future work should explore unsupervised techniques that can still capture diversity in the distribution of anomaly scores~\cite{DBLP:journals/corr/abs-2312-01488, DBLP:journals/access/KomadinaMGM24, DBLP:conf/gamesec/GhafouriALVK16, DBLP:journals/tissec/AliAKK13, DBLP:conf/kdd/SifferFTL17}.

\section*{Acknowledgements}
This work has been partially funded by (i) Project SERICS (PE00000014) under the MUR National Recovery and Resilience Plan funded by the European Union -- NextGenerationEU; (ii) MUR on D.M. 352/2022, PNRR Ricerca, CUP H23C22000550005; (iii) Moneying-Plus project (CUP J29I24001790005) selected within the framework of the PR FESR – FSE Calabria 2021/2027 and implemented with the support of the Italian State and the Calabria Region – Action 1.1.1: Support for research, development, and innovation projects, including those carried out in collaboration with research organizations, within the priority areas and trajectories of the S3 Strategy.

\bibliography{ref}
\bibliographystyle{ACM-Reference-Format}

\clearpage

\appendix
\section{Training Details and Sensitivity Analysis}
GNN hyperparameters (layers, neurons, lr) are tuned to ensure meaningful node embeddings, thus the chosen configuration minimizes the loss. After integrating the GNN with HST, we perform a second tuning phase. The list of hyperparameters is shown in Table~\ref{tab:hyperparams}.

\begin{table}[h]
\centering
\begin{tabular}{@{}ll@{}}
\toprule
\multicolumn{2}{c}{\textbf{GNN Training}} \\
\midrule
Layers         & 2--6 \\
Channels       & Hidden: 16; Output: 4, 8, 16 \\
Learning Rate  & 0.001 \\
Sampling       & Uniform \\
Aggregation    & Mean \\
\addlinespace
\toprule
\multicolumn{2}{c}{\textbf{Anomaly Scoring}} \\
\midrule
Edge Embedding & Eq.~\ref{eq:edgemean}, Eq.~\ref{eq:edgesub} \\
Weights        & (1.0, 0.0, 0.0), (0.33, 0.33, 0.33) \\
Cache Size     & 64, 32, 16, 8, 4 \\
Trees          & 8, 16, 32, 64 \\
Depth          & 3, 6, 9, 12 \\
HST Window     & 8, 64, 512, 1024, 2048, 4096, 8192, $10^5$, $10^6$ \\
\bottomrule
\end{tabular}
\caption{Hyperparameters and settings used for GNN training and anomaly scoring.}
\label{tab:hyperparams}
\end{table}

Further, Table~\ref{tab:appendix_datasets} reports some statistics about the seven real-world datasets used for the experiments
\begin{table}[ht]
\centering
\resizebox{0.5\columnwidth}{!}
{
\begin{tabular}{@{}cccc@{}}\toprule
Dataset & $|V|$ & $|E|$ & \% anomalies\\ \midrule
 DARPA & $25$K & $4.5$M & $59.9\%$ \\
 UNSW-NB15 & $50$ & $2.5$M & $12.8\%$ \\ 
 ISCX2012 & $31$K & $1.1$M & $4.2\%$ \\ 
 CIC-IDS2017 & $33$K & $7.8$M & $7.4\%$ \\ 
 CTU-13 Scenario 1 & $607$K & $2.8$M & $1.5\%$ \\ 
 CTU-13 Scenario 10 & $198$K & $1.3$M & $8.1\%$ \\ 
 CTU-13 Scenario 13 & $316$K & $1.9$M & $2.1\%$ \\ 
\bottomrule
\end{tabular}
}
\caption{Summary statistics on the datasets.}
\label{tab:appendix_datasets}
\end{table}

In Tables~\ref{tab:trees_sensitivity} and~\ref{tab:depth_sensitivity} we report the AUC and AP by varying the number of trees and the depth for each tree, respectively. Generally, more trees and more depth improve robustness by reducing variance, but the optimal number is dataset-dependent.

Further, weights in Eq.~\ref{eq:anomalyscore} and the edge embedding strategy (Eq.~\ref{eq:edgemean} and Eq.~\ref{eq:edgesub}) are also part of the hyperparameter tuning process. To provide further clarity, we include an experiment analyzing their impact on model performance in Tables~\ref{tab:weights_sensitivity} and~\ref{tab:edge_embeddings_sensitivity}.

\begin{table*}[h]
\resizebox{\textwidth}{!}{
\begin{tabular}{@{}ccccccccccccccc@{}}
\toprule
\multirow{3}{*}{Trees} & \multicolumn{2}{c}{\multirow{2}{*}{DARPA}} & \multicolumn{2}{c}{\multirow{2}{*}{UNSW-NB15}} & \multicolumn{2}{c}{\multirow{2}{*}{ISCX2012}} & \multicolumn{2}{c}{\multirow{2}{*}{CIC-IDS2017}} & \multicolumn{6}{c}{CTU-13} \\ \cmidrule(l){10-15} 
 & \multicolumn{2}{c}{} & \multicolumn{2}{c}{} & \multicolumn{2}{c}{} & \multicolumn{2}{c}{} & \multicolumn{2}{c}{Scenario 1} & \multicolumn{2}{c}{Scenario 10} & \multicolumn{2}{c}{Scenario 13} \\ \cmidrule(l){2-15} 
 & ROC-AUC & \multicolumn{1}{c|}{AP} & ROC-AUC & \multicolumn{1}{c|}{AP} & ROC-AUC & \multicolumn{1}{c|}{AP} & ROC-AUC & \multicolumn{1}{c|}{AP} & ROC-AUC & \multicolumn{1}{c|}{AP} & ROC-AUC & \multicolumn{1}{c|}{AP} & ROC-AUC & AP \\ \midrule
8 & $0.985 \pm 0.008$ & $0.991 \pm 0.005$ & $0.985 \pm 0.002$ & $0.891 \pm 0.014$ & $0.992 \pm 0.009$ & $0.701 \pm 0.247$ & $0.984 \pm 0.001$ & $0.806 \pm 0.025$ & $0.969 \pm 0.004$ & $0.290 \pm 0.015$ & $0.938 \pm 0.029$ & $0.661 \pm 0.101$ & $0.668 \pm 0.267$ & $0.118 \pm 0.126$ \\
16 & $0.988 \pm 0.003$ & $0.993 \pm 0.001$ & $0.983 \pm 0.001$ & $0.883 \pm 0.011$ & $0.996 \pm 0.005$ & $0.833 \pm 0.189$ & $0.985 \pm 0.001$ & $0.804 \pm 0.031$ & $0.968 \pm 0.004$ & $0.280 \pm 0.028$ & $0.958 \pm 0.020$ & $0.728 \pm 0.099$ & $0.794 \pm 0.216$ & $0.161 \pm 0.139$ \\
32 & $0.990 \pm 0.002$ & $0.994 \pm 0.001$ & $0.983 \pm 0.001$ & $0.883 \pm 0.011$ & $0.999 \pm 0.001$ & $0.935 \pm 0.029$ & $0.985 \pm 0.001$ & $0.781 \pm 0.038$ & $0.967 \pm 0.001$ & $0.246 \pm 0.023$ & $0.962 \pm 0.011$ & $0.659 \pm 0.069$ & $0.859 \pm 0.185$ & $0.173 \pm 0.105$ \\
64 & $0.991 \pm 0.000$ & $0.994 \pm 0.000$ & $0.984 \pm 0.001$ & $0.889 \pm 0.009$ & $0.999 \pm 0.001$ & $0.944 \pm 0.007$ & $0.985 \pm 0.001$ & $0.766 \pm 0.033$ & $0.966 \pm 0.000$ & $0.228 \pm 0.021$ & $0.961 \pm 0.007$ & $0.640 \pm 0.033$ & $0.925 \pm 0.116$ & $0.244 \pm 0.107$ \\ \bottomrule
\end{tabular}
}
\caption{Sensitivity analysis by varying the number of trees.}
\label{tab:trees_sensitivity}
\end{table*}

\begin{table*}[h]
\resizebox{\textwidth}{!}{
\centering
\begin{tabular}{@{}ccccccccccccccc@{}}
\toprule
\multirow{3}{*}{Depths} & \multicolumn{2}{c}{\multirow{2}{*}{DARPA}} & \multicolumn{2}{c}{\multirow{2}{*}{UNSW-NB15}} & \multicolumn{2}{c}{\multirow{2}{*}{ISCX2012}} & \multicolumn{2}{c}{\multirow{2}{*}{CIC-IDS2017}} & \multicolumn{6}{c}{CTU-13} \\ \cmidrule(l){10-15} 
 & \multicolumn{2}{c}{} & \multicolumn{2}{c}{} & \multicolumn{2}{c}{} & \multicolumn{2}{c}{} & \multicolumn{2}{c}{Scenario 1} & \multicolumn{2}{c}{Scenario 10} & \multicolumn{2}{c}{Scenario 13} \\ \cmidrule(l){2-15} 
 & ROC-AUC & \multicolumn{1}{c|}{AP} & ROC-AUC & \multicolumn{1}{c|}{AP} & ROC-AUC & \multicolumn{1}{c|}{AP} & ROC-AUC & \multicolumn{1}{c|}{AP} & ROC-AUC & \multicolumn{1}{c|}{AP} & ROC-AUC & \multicolumn{1}{c|}{AP} & ROC-AUC & AP \\ 
\midrule
3 & $0.985 \pm 0.008$ & $0.991 \pm 0.005$ & $0.984 \pm 0.003$ & $0.889 \pm 0.018$ & $0.999 \pm 0.000$ & $0.944 \pm 0.007$ & $0.984 \pm 0.001$ & $0.806 \pm 0.025$ & $0.966 \pm 0.004$ & $0.260 \pm 0.024$ & $0.941 \pm 0.076$ & $0.752 \pm 0.225$ & $0.925 \pm 0.116$ & $0.244 \pm 0.107$ \\
6 & $0.986 \pm 0.004$ & $0.992 \pm 0.002$ & $0.983 \pm 0.002$ & $0.882 \pm 0.012$ & $0.999 \pm 0.000$ & $0.949 \pm 0.006$ & $0.985 \pm 0.001$ & $0.810 \pm 0.027$ & $0.969 \pm 0.004$ & $0.290 \pm 0.015$ & $0.942 \pm 0.040$ & $0.660 \pm 0.133$ & $0.637 \pm 0.188$ & $0.064 \pm 0.075$ \\
9 & $0.987 \pm 0.004$ & $0.993 \pm 0.002$ & $0.984 \pm 0.002$ & $0.880 \pm 0.016$ & $0.999 \pm 0.000$ & $0.951 \pm 0.007$ & $0.984 \pm 0.001$ & $0.817 \pm 0.011$ & $0.970 \pm 0.006$ & $0.292 \pm 0.037$ & $0.957 \pm 0.020$ & $0.712 \pm 0.094$ & $0.503 \pm 0.011$ & $0.027 \pm 0.003$ \\
12 & $0.988 \pm 0.005$ & $0.994 \pm 0.002$ & $0.985 \pm 0.002$ & $0.891 \pm 0.014$ & $0.999 \pm 0.000$ & $0.951 \pm 0.004$ & $0.985 \pm 0.001$ & $0.805 \pm 0.032$ & $0.970 \pm 0.004$ & $0.290 \pm 0.027$ & $0.949 \pm 0.023$ & $0.659 \pm 0.131$ & $0.499 \pm 0.013$ & $0.024 \pm 0.003$ \\
\bottomrule
\end{tabular}
}
\caption{Sensitivity analysis by varying the tree depth.}
\label{tab:depth_sensitivity}
\end{table*}

\begin{table*}[h]
\resizebox{\textwidth}{!}{
\begin{tabular}{@{}ccccccccccccccc@{}}
\toprule
\multirow{3}{*}{Weights} & \multicolumn{2}{c}{\multirow{2}{*}{DARPA}} & \multicolumn{2}{c}{\multirow{2}{*}{UNSW-NB15}} & \multicolumn{2}{c}{\multirow{2}{*}{ISCX2012}} & \multicolumn{2}{c}{\multirow{2}{*}{CIC-IDS2017}} & \multicolumn{6}{c}{CTU-13} \\ \cmidrule(l){10-15} 
 & \multicolumn{2}{c}{} & \multicolumn{2}{c}{} & \multicolumn{2}{c}{} & \multicolumn{2}{c}{} & \multicolumn{2}{c}{Scenario 1} & \multicolumn{2}{c}{Scenario 10} & \multicolumn{2}{c}{Scenario 13} \\ \cmidrule(l){2-15} 
 & ROC-AUC & \multicolumn{1}{c|}{AP} & ROC-AUC & \multicolumn{1}{c|}{AP} & ROC-AUC & \multicolumn{1}{c|}{AP} & ROC-AUC & \multicolumn{1}{c|}{AP} & ROC-AUC & \multicolumn{1}{c|}{AP} & ROC-AUC & \multicolumn{1}{c|}{AP} & ROC-AUC & AP \\ \midrule
0.0, 0.0, 1.0 & $0.985 \pm 0.008$ & $0.991 \pm 0.005$ & $0.985 \pm 0.002$ & $0.892 \pm 0.014$ & $0.999 \pm 0.000$ & $0.942 \pm 0.007$ & $0.984 \pm 0.001$ & $0.806 \pm 0.025$ & $0.969 \pm 0.004$ & $0.291 \pm 0.014$ & $0.956 \pm 0.019$ & $0.712 \pm 0.086$ & $0.948 \pm 0.049$ & $0.249 \pm 0.091$ \\
0.33, 0.33, 0.33 & $0.977 \pm 0.007$ & $0.991 \pm 0.003$ & $0.657 \pm 0.255$ & $0.433 \pm 0.242$ & $0.931 \pm 0.005$ & $0.255 \pm 0.012$ & $0.966 \pm 0.080$ & $0.856 \pm 0.244$ & $0.964 \pm 0.003$ & $0.267 \pm 0.015$ & $0.985 \pm 0.004$ & $0.787 \pm 0.047$ & $0.520 \pm 0.005$ & $0.023 \pm 0.004$ \\ \bottomrule
\end{tabular}
}
\caption{Sensitivity analysis by varying nodes and edge weights for anomaly scoring.}
\label{tab:weights_sensitivity}
\end{table*}

\begin{table*}[h]
\resizebox{\textwidth}{!}{
\begin{tabular}{@{}ccccccccccccccc@{}}
\toprule
\multirow{3}{*}{Edge Embeddings} & \multicolumn{2}{c}{\multirow{2}{*}{DARPA}} & \multicolumn{2}{c}{\multirow{2}{*}{UNSW-NB15}} & \multicolumn{2}{c}{\multirow{2}{*}{ISCX2012}} & \multicolumn{2}{c}{\multirow{2}{*}{CIC-IDS2017}} & \multicolumn{6}{c}{CTU-13} \\ \cmidrule(l){10-15} 
 & \multicolumn{2}{c}{} & \multicolumn{2}{c}{} & \multicolumn{2}{c}{} & \multicolumn{2}{c}{} & \multicolumn{2}{c}{Scenario 1} & \multicolumn{2}{c}{Scenario 10} & \multicolumn{2}{c}{Scenario 13} \\ \cmidrule(l){2-15} 
 & ROC-AUC & \multicolumn{1}{c|}{AP} & ROC-AUC & \multicolumn{1}{c|}{AP} & ROC-AUC & \multicolumn{1}{c|}{AP} & ROC-AUC & \multicolumn{1}{c|}{AP} & ROC-AUC & \multicolumn{1}{c|}{AP} & ROC-AUC & \multicolumn{1}{c|}{AP} & ROC-AUC & AP \\ \midrule
Eq.1 & $0.985 \pm 0.008$ & $0.991 \pm 0.005$ & $0.109 \pm 0.004$ & $0.149 \pm 0.005$ & $0.431 \pm 0.270$ & $0.116 \pm 0.210$ & $0.984 \pm 0.001$ & $0.806 \pm 0.025$ & $0.969 \pm 0.004$ & $0.290 \pm 0.015$ & $0.770 \pm 0.106$ & $0.391 \pm 0.110$ & $0.469 \pm 0.014$ & $0.023 \pm 0.003$ \\
Eq.2 & $0.844 \pm 0.048$ & $0.866 \pm 0.055$ & $0.985 \pm 0.002$ & $0.891 \pm 0.014$ & $0.999 \pm 0.000$ & $0.942 \pm 0.007$ & $0.981 \pm 0.004$ & $0.777 \pm 0.038$ & $0.886 \pm 0.052$ & $0.084 \pm 0.025$ & $0.957 \pm 0.020$ & $0.712 \pm 0.094$ & $0.925 \pm 0.116$ & $0.244 \pm 0.107$ \\ \bottomrule
\end{tabular}
}
\caption{Sensitivity analysis by varying the edge embedding strategy.}
\label{tab:edge_embeddings_sensitivity}
\end{table*}

\section{Further Experiments}

To strengthen and broaden our experimental evaluation, we provide two additional experiments using non-cybersecurity related datasets, namely Bitcoin-Alpha and Bitcoin-OTC. These datasets represent financial trust networks, where nodes correspond to users and edges indicate trust scores assigned from one user to another. In this setting, abnormality is determined by the likelihood that a user is suspicious based on the proportion of low trust scores they receive. This approach differs significantly from the scenarios considered in our experiments, which are rooted in cybersecurity contexts where anomalies come from malicious activities such as cyber-attacks.
For consistency and comparison, we adopted the same experimental setup as SLADE, including temporal splits and abnormality labeling, allowing us to directly reference their reported results.

In Table~\ref{tab:bitcoin_experiment}, we present a comparative analysis of \themodel (in both static and dynamic configurations) against SLADE and other baseline methods such as AnoEdge-l and MIDAS.
SLADE consistently achieves the highest AUC, while the proposed model outperforms all baselines, including SLADE, in terms of average precision (AP). These additional experiments highlight the potentiality and generality of the proposed method.

We also assess the statistical robustness of our framework by evaluating its performance under varying data conditions. Table~\ref{tab:appendix_statistical_robustness} reports the ROC-AUC and AP scores for \themodel and baseline models across several datasets, where we uniformly sample $\%50$ of the test set. Notably, ARES demonstrates greater stability, maintaining high detection performance with minimal degradation, whereas baseline methods often exhibit larger performance drops or increased variability. This highlights the robustness of ARES not only in full-data scenarios but also under constrained or noisy data conditions.

\begin{table}[ht]
\centering
\resizebox{0.6\columnwidth}{!}{
\begin{tabular}{@{}ccccc@{}}
\toprule
\multirow{3}{*}{Model} & \multicolumn{2}{c}{\multirow{2}{*}{Bitcoin Alpha}} & \multicolumn{2}{c}{\multirow{2}{*}{Bitcoin OTC}} \\
 & \multicolumn{2}{c}{} & \multicolumn{2}{c}{} \\ \cmidrule(l){2-5} 
 & ROC-AUC & \multicolumn{1}{c|}{AP} & ROC-AUC & AP \\ \midrule
MIDAS-F & $0.646 \pm 0.011$ & $0.131 \pm 0.001$ & $0.622 \pm 0.021$ & $0.161 \pm 0.000$ \\
AnoEdge-L & $0.625 \pm 0.002$ & $0.125 \pm 0.011$ & $0.661 \pm 0.19$ & $0.170 \pm 0.001$ \\
SLADE-H & \bm{$0.769 \pm 0.004$} & $0.149 \pm 0.001$ & \bm{$0.772 \pm 0.003$} & $0.205 \pm 0.001$ \\ \midrule
ARES-Static & $0.729 \pm 0.004$ & $0.161 \pm 0.003$ & $0.757 \pm 0.013$ & \bm{$0.264 \pm 0.026$} \\
ARES-Dynamic & $0.731 \pm 0.004$ & \bm{$0.167 \pm 0.007$} & $0.592 \pm 0.040$ & $0.111 \pm 0.012$ \\ \bottomrule
\end{tabular}
}
\caption{Performance comparison between \themodel and baseline methods on the Bitcoin Alpha and Bitcoin OTC datasets.}
\label{tab:bitcoin_experiment}
\end{table}

\begin{table*}[h]
\resizebox{\textwidth}{!}{
\begin{tabular}{@{}ccccccccccccccc@{}}
\toprule
\multirow{3}{*}{Model} & \multicolumn{2}{c}{\multirow{2}{*}{DARPA}} & \multicolumn{2}{c}{\multirow{2}{*}{UNSW-NB15}} & \multicolumn{2}{c}{\multirow{2}{*}{ISCX2012}} & \multicolumn{2}{c}{\multirow{2}{*}{CIC-IDS2017}} & \multicolumn{6}{c}{CTU-13} \\ \cmidrule(l){10-15} 
 & \multicolumn{2}{c}{} & \multicolumn{2}{c}{} & \multicolumn{2}{c}{} & \multicolumn{2}{c}{} & \multicolumn{2}{c}{Scenario 1} & \multicolumn{2}{c}{Scenario 10} & \multicolumn{2}{c}{Scenario 13} \\ \cmidrule(l){2-15} 
 & ROC-AUC & \multicolumn{1}{c|}{AP} & ROC-AUC & \multicolumn{1}{c|}{AP} & ROC-AUC & \multicolumn{1}{c|}{AP} & ROC-AUC & \multicolumn{1}{c|}{AP} & ROC-AUC & \multicolumn{1}{c|}{AP} & ROC-AUC & \multicolumn{1}{c|}{AP} & ROC-AUC & AP \\ \midrule
MIDAS-F & 
$0.961 \pm 0.005$ & $0.979 \pm 0.003$ & 
$0.793 \pm 0.003$ & $0.337 \pm 0.004$ & 
$0.974 \pm 0.004$ & $0.287 \pm 0.031$ & 
\bm{$0.997 \pm 0.000$} & \bm{$0.969 \pm 0.014$} & 
$0.835 \pm 0.037$ & $0.077 \pm 0.014$ & 
\bm{$0.924 \pm 0.020$} & $0.530 \pm 0.078$ & 
\bm{$0.934 \pm 0.050$} & $0.135 \pm 0.052$ \\
AnoEdge-L & 
$0.910 \pm 0.007$ & $0.955 \pm 0.003$ & 
$0.750 \pm 0.021$ & $0.443 \pm 0.033$ & 
$0.869 \pm 0.038$ & $0.097 \pm 0.027$ & 
$0.954 \pm 0.006$ & $0.486 \pm 0.028$ & 
$0.479 \pm 0.007$ & $0.018 \pm 0.001$ & 
$0.692 \pm 0.006$ & $0.181 \pm 0.010$ & 
$0.293 \pm 0.009$ & $0.010 \pm 0.000$ \\
SLADE-H &  
$0.900 \pm 0.093$ & \bm{$0.912\pm 0.091$} & 
\bm{$0.911 \pm 0.118$} & \bm{$0.757 \pm 0.191$} &
\bm{$0.999 \pm 0.000$} & \bm{$0.985 \pm 0.022$} &
\bm{$0.731 \pm 0.226$} & $0.215 \pm 0.178$ &
$0.564 \pm 0.077$ & $0.018 \pm 0.005$ &
$0.399 \pm 0.159$ & $0.099 \pm 0.020$ &
$0.564 \pm 0.077$ & $0.018 \pm 0.005$  
\\ \midrule
\themodel-Static & 
\bm{$0.991 \pm 0.005$} & \bm{$0.994 \pm 0.003$} & 
\bm{$0.984 \pm 0.003$} & \bm{$0.885 \pm 0.016$} & 
\bm{$0.999 \pm 0.000$} & $0.945 \pm 0.010$ & 
$0.984 \pm 0.001$ & $0.811 \pm 0.017$ & 
\bm{$0.970 \pm 0.003$} & \bm{$0.296 \pm 0.030$} & 
\bm{$0.945 \pm 0.024$} & \bm{$0.701 \pm 0.088$} & 
\bm{$0.928 \pm 0.122$} & \bm{$0.282 \pm 0.119$} \\
\themodel-Dynamic & 
\bm{$0.989 \pm 0.007$} & \bm{$0.994 \pm 0.003$} & 
$0.877 \pm 0.009$ & $0.500 \pm 0.030$ & 
\bm{$0.999 \pm 0.000$} & \bm{$0.960 \pm 0.018$} & 
$0.984 \pm 0.001$ & $0.816 \pm 0.013$ & 
\bm{$0.947 \pm 0.027$} & \bm{$0.209 \pm 0.073$} & 
\bm{$0.924 \pm 0.022$} & \bm{$0.673 \pm 0.091$} & 
\bm{$0.851 \pm 0.042$} & \bm{$0.098 \pm 0.040$} \\ \bottomrule
\end{tabular}
}
\caption{Comparison of \themodel against competing methods by evaluating performance on subsampled versions of the test set for each dataset.}
\label{tab:appendix_statistical_robustness}
\end{table*}

\section{Ablation Study}\label{sec:appendix_ablation}

\begin{table*}[htbp]
\resizebox{\textwidth}{!}{
\begin{tabular}{@{}ccccccccccccccc@{}}
\toprule
\multirow{3}{*}{Model} & \multicolumn{2}{c}{\multirow{2}{*}{DARPA}} & \multicolumn{2}{c}{\multirow{2}{*}{UNSW-NB15}} & \multicolumn{2}{c}{\multirow{2}{*}{ISCX2012}} & \multicolumn{2}{c}{\multirow{2}{*}{CIC-IDS2017}} & \multicolumn{6}{c}{CTU-13} \\ \cmidrule(l){10-15} 
 & \multicolumn{2}{c}{} & \multicolumn{2}{c}{} & \multicolumn{2}{c}{} & \multicolumn{2}{c}{} & \multicolumn{2}{c}{Scenario 1} & \multicolumn{2}{c}{Scenario 10} & \multicolumn{2}{c}{Scenario 13} \\ \cmidrule(l){2-15} 
 & ROC-AUC & \multicolumn{1}{c|}{AP} & ROC-AUC & \multicolumn{1}{c|}{AP} & ROC-AUC & \multicolumn{1}{c|}{AP} & ROC-AUC & \multicolumn{1}{c|}{AP} & ROC-AUC & \multicolumn{1}{c|}{AP} & ROC-AUC & \multicolumn{1}{c|}{AP} & ROC-AUC & AP \\ \midrule
GNN-RecErr & $0.304 \pm 0.000$ & $0.034 \pm 0.000$ & $0.629 \pm 0.000$ & $0.247 \pm 0.000$ & $0.559 \pm 0.000$ & $0.059 \pm 0.000$ & $0.419 \pm 0.000$ & $0.126 \pm 0.000$ & $0.394 \pm 0.000$ & $0.022 \pm 0.000$ & $0.222 \pm 0.000$ & $0.041 \pm 0.000$ & $0.487 \pm 0.000$ & $0.023 \pm 0.000$ \\ \bottomrule
GCN+HST & $0.637 \pm 0.101$ & $0.806 \pm 0.039$ & \bm{$0.986 \pm 0.001$} & \bm{$0.894 \pm 0.004$} & $0.969 \pm 0.009$ & $0.347 \pm 0.085$ & $0.881 \pm 0.061$ & $0.352 \pm 0.096$ & $0.707 \pm 0.173$ & $0.066 \pm 0.032$ & $0.348 \pm 0.071$ & $0.138 \pm 0.010$ & $0.817 \pm 0.059$ & $0.068 \pm 0.019$ \\
GAT+HST & $0.965 \pm 0.004$ & $0.975 \pm 0.003$ & $0.688 \pm 0.296$ & $0.521 \pm 0.289$ & $0.839 \pm 0.072$ & $0.138 \pm 0.058$ & \bm{$0.974 \pm 0.024$} & \bm{$0.734 \pm 0.140$} & $0.654 \pm 0.009$ & $0.034 \pm 0.005$ & \bm{$0.914 \pm 0.076$} & \bm{$0.592 \pm 0.194$} & $0.791 \pm 0.002$ & $0.037 \pm 0.001$  \\ \bottomrule
GraphSage+RRCF & \bm{$0.947 \pm 0.041$} & \bm{$0.955 \pm 0.035$} & $0.811 \pm 0.059$ & $0.538 \pm 0.069$ & $0.995 \pm 0.001$ & $0.679 \pm 0.011$ & $0.616 \pm 0.097$ & $0.347 \pm 0.059$ & $0.804 \pm 0.064$ & $0.072 \pm 0.025$ & $0.728 \pm 0.041$ & $0.226 \pm 0.036$ & $0.506 \pm 0.105$ & $0.025 \pm 0.005$ \\ \bottomrule
\themodel & \bm{$0.985 \pm 0.008$} & \bm{$0.991 \pm 0.005$} & \bm{$0.985 \pm 0.002$} & \bm{$0.892 \pm 0.014$} & \bm{$0.999 \pm 0.000$} & \bm{$0.942 \pm 0.007$} & \bm{$0.984 \pm 0.001$} & \bm{$0.806 \pm 0.025$} & \bm{$0.969 \pm 0.004$} & \bm{$0.291 \pm 0.014$} & \bm{$0.956 \pm 0.019$} & \bm{$0.712 \pm 0.086$} & \bm{$0.948 \pm 0.049$} & \bm{$0.249 \pm 0.091$} \\ \bottomrule
\end{tabular}
}
\caption{Ablation study results on multiple datasets, highlighting the effectiveness of combining GraphSAGE embeddings with Half-Space Trees.}
\label{tab:ablation}
\end{table*}

\paragraph{\textbf{Half-Space Trees contribution}.} 
HST provides (i) a simple but effective way of detecting anomalies, and (ii) it adapts to concept drift by continuously updating statistics, capturing data distribution shifts. Nonetheless, HST is not directly applicable to graph data, making an isolated ablation unfeasible. The GNN, used solely for node embeddings, can reflect drift when an increase in the reconstruction error occurs, offering a potential proxy for change detection. While not directly explored in our experiments, we acknowledge this potential and plan to incorporate continual learning methods to improve adaptability in future work.

That said, we compare ARES with a GNN-only variant using reconstruction error as the anomaly score following prior work~\cite{DBLP:journals/tkde/MaWXYZSXA23} (GNN-RecErr) and an ablation study comparing HST with RRCF~\cite{DBLP:conf/icml/GuhaMRS16} (GraphSage+RRCF). Results (AUC and AP) reported in Table~\ref{tab:ablation} demonstrate ARES significantly outperforms the baselines. 

Notably, as previously mentioned, HST consistently outperforms the more recent RRCF, achieving superior accuracy while being 2x to 9x faster in inference time across multiple datasets as depicted in Table~\ref{tab:timings_hst_rrcf}. These results further validate HST as a robust and efficient choice for real-time anomaly detection in streaming graph environments.
\paragraph{\textbf{Different Node/Edge embedders.}} 
GraphSAGE has been shown to produce meaningful representations of nodes and edges that effectively capture the underlying graph structure. We conducted further experiments comparing ARES (GraphSAGE + HST) with GCN+HST and GAT+HST described in Table~\ref{tab:ablation}

\begin{table*}[hb!]
\centering
\resizebox{0.4\columnwidth}{!}{
\begin{tabular}{@{}ccc@{}}
\toprule
Dataset & ARES & RRCF \\ \midrule
DARPA & $7.8$ & $33.9$ \\
UNSW-NB15 & $5.4$ & $10.7$ \\
ISCX2012 & $14.1$ & $28.3$ \\
CIC-IDS2017 & $51.4$ & $457.6$ \\
CTU-13 Scenario 1 & $59.5$  & $260.2$\\
CTU-13 Scenario 10 & $44.9$  & $85.4$ \\
CTU-13 Scenario 13 & $108.3$  &  $203.5$\\ \bottomrule
\end{tabular}
}
\caption{Average inference time (in seconds) across multiple datasets for \themodel (HST-based) and RRCF.} \label{tab:timings_hst_rrcf}
\end{table*}

\section{Thresholding mechanism limitations}
Additional observations can also be made on the adaptive thresholding method, whose sensitivity to overfitting depends on the choice of the validation set. In our experiments, we partition the data according to the temporal dimension. In presence of dramatic concept drifts, this can result in a poor choice of the optimal thresholds, as witnessed by the unstable results in terms of F1 and B-Accuracy. For example, the validation set for ISCX2012 comprises $99.962\%$ normal edges and only $0.038\%$ anomalies, whereas the test set consists of $97.6\%$ normal edges and $2.4\%$ anomalous edges. In such, computing a threshold according to a validation set which exhibits a different distribution of anomalies may result in unstable results. 

\end{document}